\PassOptionsToPackage{unicode}{hyperref}
\PassOptionsToPackage{hyphens}{url}
\PassOptionsToPackage{usenames,dvipsnames,svgnames,x11names}{xcolor}
\documentclass[
  12pt]{article}

\usepackage{amsmath,amssymb}
\usepackage{iftex}
\ifPDFTeX
  \usepackage[T1]{fontenc}
  \usepackage[utf8]{inputenc}
  \usepackage{textcomp} 
\else 
  \usepackage{unicode-math}
  \defaultfontfeatures{Scale=MatchLowercase}
  \defaultfontfeatures[\rmfamily]{Ligatures=TeX,Scale=1}
\fi
\usepackage{lmodern}
\ifPDFTeX\else  
\fi
\IfFileExists{upquote.sty}{\usepackage{upquote}}{}
\IfFileExists{microtype.sty}{
  \usepackage[]{microtype}
  \UseMicrotypeSet[protrusion]{basicmath} 
}{}
\makeatletter
\@ifundefined{KOMAClassName}{
  \IfFileExists{parskip.sty}{%
    \usepackage{parskip}
  }{
    \setlength{\parindent}{0pt}
    \setlength{\parskip}{6pt plus 2pt minus 1pt}}
}{
  \KOMAoptions{parskip=half}}
\makeatother
\usepackage{xcolor}
\makeatletter
\ifx\paragraph\undefined\else
  \let\oldparagraph\paragraph
  \renewcommand{\paragraph}{
    \@ifstar
      \xxxParagraphStar
      \xxxParagraphNoStar
  }
  \newcommand{\xxxParagraphStar}[1]{\oldparagraph*{#1}\mbox{}}
  \newcommand{\xxxParagraphNoStar}[1]{\oldparagraph{#1}\mbox{}}
\fi
\ifx\subparagraph\undefined\else
  \let\oldsubparagraph\subparagraph
  \renewcommand{\subparagraph}{
    \@ifstar
      \xxxSubParagraphStar
      \xxxSubParagraphNoStar
  }
  \newcommand{\xxxSubParagraphStar}[1]{\oldsubparagraph*{#1}\mbox{}}
  \newcommand{\xxxSubParagraphNoStar}[1]{\oldsubparagraph{#1}\mbox{}}
\fi
\makeatother

\usepackage{longtable,booktabs,array}
\usepackage{calc} 
\usepackage{etoolbox}
\makeatletter
\patchcmd\longtable{\par}{\if@noskipsec\mbox{}\fi\par}{}{}
\makeatother
\IfFileExists{footnotehyper.sty}{\usepackage{footnotehyper}}{\usepackage{footnote}}
\makesavenoteenv{longtable}
\usepackage{graphicx}
\makeatletter
\def\maxwidth{\ifdim\Gin@nat@width>\linewidth\linewidth\else\Gin@nat@width\fi}
\def\maxheight{\ifdim\Gin@nat@height>\textheight\textheight\else\Gin@nat@height\fi}
\makeatother
\setkeys{Gin}{width=\maxwidth,height=\maxheight,keepaspectratio}
\makeatletter
\def\fps@figure{htbp}
\makeatother

\makeatletter
\@ifpackageloaded{caption}{}{\usepackage{caption}}
\AtBeginDocument{%
\ifdefined\contentsname
  \renewcommand*\contentsname{Table of contents}
\else
  \newcommand\contentsname{Table of contents}
\fi
\ifdefined\listfigurename
  \renewcommand*\listfigurename{List of Figures}
\else
  \newcommand\listfigurename{List of Figures}
\fi
\ifdefined\listtablename
  \renewcommand*\listtablename{List of Tables}
\else
  \newcommand\listtablename{List of Tables}
\fi
\ifdefined\figurename
  \renewcommand*\figurename{Figure}
\else
  \newcommand\figurename{Figure}
\fi
\ifdefined\tablename
  \renewcommand*\tablename{Table}
\else
  \newcommand\tablename{Table}
\fi
}
\@ifpackageloaded{float}{}{\usepackage{float}}
\floatstyle{ruled}
\@ifundefined{c@chapter}{\newfloat{codelisting}{h}{lop}}{\newfloat{codelisting}{h}{lop}[chapter]}
\floatname{codelisting}{Listing}

\makeatother
\makeatletter
\@ifpackageloaded{caption}{}{\usepackage{caption}}
\@ifpackageloaded{subcaption}{}{\usepackage{subcaption}}
\makeatother

\ifLuaTeX
  \usepackage{selnolig}  
\fi
\usepackage{natbib}
\usepackage{bookmark}

\IfFileExists{xurl.sty}{\usepackage{xurl}}{} 
\hypersetup{
  pdftitle={Title},
  pdfauthor={Author 1; Author 2},
  pdfkeywords={3 to 6 keywords, that do not appear in the title},
  colorlinks=true,
  linkcolor={blue},
  filecolor={Maroon},
  citecolor={Blue},
  urlcolor={Blue},
  pdfcreator={LaTeX via pandoc}}

\usepackage[T1]{fontenc}
\usepackage{amsmath}
\usepackage{amsfonts}
\usepackage{amscd}
\usepackage{amssymb}
\usepackage{amsthm}
\usepackage{bm}
\usepackage{thmtools}
\usepackage{thm-restate}

\usepackage{enumerate}
\usepackage{enumitem}

\usepackage{epsfig}

\usepackage{url}

\usepackage{booktabs,multirow,multicol}       

\usepackage[nameinlink]{cleveref}
\creflabelformat{equation}{#1#2#3}
\crefname{equation}{eq.}{eqs.}
\Crefname{equation}{Eq.}{Eqs.}
\Crefname{section}{\S}{\S}

\DeclareRobustCommand{\parhead}[1]{\textbf{#1}~}

\numberwithin{equation}{section}

\usepackage{caption}
\usepackage{subcaption}
\newtheorem{theorem}{Theorem}

\newtheorem{definition}{Definition}
\newtheorem{lemma}{Lemma}

\newtheorem{remark}{Remark}

\newtheorem{assumption}{Assumption}

\crefname{assumption}{assumption}{assumptions}

\newcommand{\defeq}{\mathrel{\mathop:}=}

\newcommand{\indep}{\perp \!\!\! \perp}

\usepackage{tcolorbox}

\newcommand{\zpr}{p_{\bm{z}}}

\usepackage{amsmath,amsfonts,bm}

\def\1{\bm{1}}

\DeclareMathAlphabet{\mathsfit}{\encodingdefault}{\sfdefault}{m}{sl}
\SetMathAlphabet{\mathsfit}{bold}{\encodingdefault}{\sfdefault}{bx}{n}

\usepackage[algoruled,ruled,vlined]{algorithm2e}
\usepackage{listings}
\usepackage{fancyvrb}
\fvset{fontsize=\normalsize}

\makeatletter
\renewcommand{\SetKwInOut}[2]{%
  \sbox\algocf@inoutbox{\KwSty{#2}\algocf@typo:}%
  \expandafter\ifx\csname InOutSizeDefined\endcsname\relax
    \newcommand\InOutSizeDefined{}\setlength{\inoutsize}{\wd\algocf@inoutbox}%
    \sbox\algocf@inoutbox{\parbox[t]{\inoutsize}{\KwSty{#2}\algocf@typo:\hfill}~}\setlength{\inoutindent}{\wd\algocf@inoutbox}%
  \else
    \ifdim\wd\algocf@inoutbox>\inoutsize%
    \setlength{\inoutsize}{\wd\algocf@inoutbox}%
    \sbox\algocf@inoutbox{\parbox[t]{\inoutsize}{\KwSty{#2}\algocf@typo:\hfill}~}\setlength{\inoutindent}{\wd\algocf@inoutbox}%
    \fi%
  \fi
  \algocf@newcommand{#1}[1]{%
    \ifthenelse{\boolean{algocf@inoutnumbered}}{\relax}{\everypar={\relax}}%
    {\let\\\algocf@newinout\hangindent=\inoutindent\hangafter=1\parbox[t]{\inoutsize}{\KwSty{#2}\algocf@typo:\hfill}~##1\par}%
    \algocf@linesnumbered
  }}%
\makeatother

\SetKwInOut{KwInput}{input}
\SetKwInOut{KwOutput}{output}

\newcommand{\disablecomments}{
    \newcommand{\gemma}[1]{}
    \newcommand{\hilite}[1]{\textcolor{red}{##1}}
}

\disablecomments

\usepackage[affil-it]{authblk}

\title{Nonlinear multi-study sparse factor analysis}
\date{}          

\author[1]{Gemma E. Moran}
\author[2,3]{Anandi Krishnan}
\affil[1]{Department of Statistics, Rutgers University}
\affil[2]{Department of Biomedical Engineering, Rutgers University}
\affil[3]{School of Medicine, Stanford University}

\begin{document}

\def\spacingset#1{\renewcommand{\baselinestretch}%
{#1}\small\normalsize} \spacingset{1}


\maketitle

\begin{abstract}
High-dimensional data often exhibit variation that can be captured by lower-dimensional factors. For high-dimensional data from multiple studies, one goal is to understand which underlying factors are common to all studies, and which factors are study-specific. As a particular example, we consider platelet gene expression data from patients in different disease groups. In this data, factors correspond to clusters of genes which are co-expressed; we may expect some clusters (or biological pathways) to be active for all diseases, while some clusters are only active for a specific disease. To learn these factors, we consider a nonlinear multi-study sparse factor model, which allows for both shared and study-specific factors.  To fit this model, we propose a multi-study sparse variational autoencoder. The underlying model is sparse in that each observed feature (i.e. each dimension of the data) depends on a small subset of the latent factors.  In the genomics example, this means each gene is active in only a few biological processes. We prove that the latent factor distributions are identifiable (up to element-wise transformations), and that the canonical shared and study-specific support structure is identifiable. Empirically, we demonstrate our method recovers meaningful factors in the platelet gene expression data.
\end{abstract}

\spacingset{1.2}


\section{Introduction}

In many domains, high-dimensional data have a much smaller intrinsic dimensionality. A common goal is to reduce such high-dimensional data to  lower-dimensional latent factors.  Often, instead of a single high-dimensional dataset, we observe multiple datasets from different studies or environments. A key question with such data is: what factors are shared across studies, and what factors are study-specific? In genomics, for example, it is common to collect gene expression measurements from patients from different hospitals, or with different conditions.

 As a particular example, we study gene expression levels in blood platelets from patients with diverse diseases. Platelets serve as a ``circulating biosensor'' of systemic signals including inflammation, immune activation, and drug effects \citep{Anandi2025,thomas_heterogeneity_2024,ROWLEY2019139,weyrich2014platelets}, with transcriptomes that exhibit both conserved and disease-specific pathway dysregulation (see \Cref{sec:platelet} for details). Identifying which genes are shared versus disease-specific can reveal shared biomarkers and disease-specific therapeutic targets.

In the statistics literature, recent work on this problem includes multi-study factor analysis \citep[MSFA,][]{de2019multi} and subspace factor analysis \citep[SUFA,][]{chandra2025inferring}. In MSFA, a linear factor model is introduced with a common loadings matrix shared across studies, and study-specific loadings matrices.  \citet{chandra2025inferring} propose an alternative parameterization of this model to identify the shared loadings from the study-specific loadings. 

In the machine learning literature, recent work models multi-study data using autoencoding frameworks, where an encoder network maps data to a low-dimensional representation and a decoder network maps back, jointly trained to capture the important information in the original data. This paradigm has been extended to multi-study data primarily by splitting the decoder into shared and study-specific components, combined with regularization \citep{davison2019cross,weinberger2022disentangling}.

Both the statistical and machine learning approaches to this problem have advantages and disadvantages.  The advantages of the statistical approach to multi-study factor analysis are (i) interpretability; (ii) identifiability guarantees and (iii) principled uncertainty quantification, especially for Bayesian methods. However, a disadvantage of the statistical approach is the lack of flexibility; only linear models are considered. Meanwhile, the machine learning approaches allow for flexible function estimation via neural networks, but lack guarantees regarding identifiability, and are often not directly interpretable. 

In this paper, we introduce a new method for multi-study factor analysis. In particular, we propose a nonlinear multi-study sparse factor model which we fit using a sparse variational autoencoder \citep[VAE,][]{kingma2013auto}. First, the model is flexible in that we use neural networks to learn the shared and study-specific latent factors. Second, the model is interpretable in that we can inspect which features (e.g. genes) depend on which latent factor dimensions; this allows us to interpret the factors in a similar way to sparse linear factor analysis. Third, we prove that we can identify the latent factors up to element-wise transformations and permutations. In particular, we can identify the shared structure from the study-specific structure.

\parhead{Overview.} In \Cref{sec:rel-work} we provide a more thorough literature review of statistical and machine learning approaches to the multi-study problem. In \Cref{sec:methods-msdgm} we introduce our method, multi-study sparse variational autoencoder (MSSVAE). In \Cref{sec:id-theory} we prove identifiability for the single-study model and extend to the multi-study model. 
In \Cref{sec:sim-study}, we demonstrate the performance of the MSSVAE on a variety of synthetic datasets.  In \Cref{sec:platelet}, we analyze the real platelet gene expression data.

\subsection{Related Work}\label{sec:rel-work}

\parhead{Linear methods.} A number of recent works propose linear factor models for the multi-study setting. \citet{de2019multi} introduced the multi-study factor analysis (MSFA) model, with a shared loadings matrix across studies and study-specific loadings matrices; \citet{de2021bayesian} extended MSFA to high-dimensional data with sparsity-inducing priors, fit via MCMC, and \citet{hansen2023fast} considered variational inference for the same model.

\citet{chandra2025inferring} highlight that linear multi-study factor models can have an identifiability problem; specifically, the solution where the shared factors are zero and the study-specific factors have all the signal is observationally indistinguishable from the model which correctly separates shared and study specific factors. To fix this identifiability problem, they propose an alternative linear parameterization. 

\citet{sturma2023unpaired} consider the linear multi-study setting where the latent factors now have an unknown dependence structure. They posit a linear model with shared and study-specific latent factors, introduce an algorithm to first identify shared variation across studies, and then learn a causal graph over the shared latent factors, with identifiability guarantees.

Finally, \citet{grabski2023bayesian} study a linear multi-study factor model where factors can be shared across a subset of the studies (instead of all the studies); these subsets are learned jointly with the latent factors. 

\parhead{Nonlinear methods.} Nonlinear approaches to the multi-study setting have also been recently proposed. \citet{davison2019cross} propose the Cross-Population Variational Autoencoder, where the decoder is a linear combination of a nonlinear function of the shared factors with separate nonlinear functions of the study-specific factors; in contrast, our approach uses a single nonlinear decoder. \citet{weinberger2022disentangling} propose a more flexible model, multiGroupVI, with shared and study-specific encoders mapped back to the data space via a single shared decoder; to find study-specific genes, they compare Bayes factors between the full model and a shared-factor-only model. Unlike multiGroupVI, our approach is directly interpretable in terms of which features are associated with shared versus study-specific factors, due to the sparsity in our decoder.

\parhead{Single-study sparse nonlinear factor analysis.} \citet{moran2022identifiable} consider a single-study nonlinear factor analysis model and prove that the latent factors are identifiable. Our model generalizes \citet{moran2022identifiable} to the multi-study case, using a different proof strategy that allows for more general factor distributions, identifies the dependence structure between data and factors, and extends to the negative-binomial likelihood setting; see \Cref{sec:id-theory} for details.
\section{Methods}

\subsection{Review: Multi-study linear factor analysis}

The observed data from study $m$ is $\bm{X}^{(m)}\in \mathbb{R}^{n_m \times G}$, $m=1,\dots, M$, where $n_m$ is the number of samples from study $m$ and $G$ is the number of observed features (e.g. genes).  Note that for each study, we observe the same set of features.

We first review the linear multi-study factor analysis (MSFA) model of \citet{de2019multi}. The MSFA model is:
\begin{align}
\bm{x}_{i}^{(m)} = \bm{\Phi}\bm{z}_{i}^{(m)} + \bm{\Lambda}^{(m)}\bm{\zeta}_i^{(m)} + \bm{\varepsilon}_i^{(m)}, \quad \bm{\varepsilon}_i^{(m)}\stackrel{iid}{\sim} N(0, \bm{\Sigma}^{(m)}),\quad i=1,\dots, n_m, \label{eq:unpaired-linear}
\end{align}
where $\bm{\Phi}\in\mathbb{R}^{G\times K_S}$ is the latent loadings matrix shared across studies, $\bm{z}_i^{(m)} \in \mathbb{R}^{K_S}$ are the latent factors in the shared subspace, $\bm{\Lambda}^{(m)} \in \mathbb{R}^{G\times K_m}$ is the  $m$th study specific latent loadings matrix and $\bm{\zeta}_i^{(m)}$ the corresponding latent factors. The noise vector $\bm{\varepsilon}$ has study-specific diagonal covariance $\bm{\Sigma}^{(m)} = \text{diag}\{\sigma_1^{(m)}, \dots, \sigma_G^{(m)}\}$. Note that the data is not paired (i.e. $\bm{x}_i^{(m)}$ and $\bm{x}_i^{(m')}$ are data from different samples).


For each column $\bm{\Phi}_{\cdot k} \in\mathbb{R}^G$, the genes with nonzero values can be interpreted as cluster. Genes within a cluster have correlated expression levels across all studies. Such gene clusters may be useful for developing hypotheses about biological processes which are shared across studies. (Note that genes can belong to more than one cluster.)  Similarly, $\bm{\Lambda}_{\cdot k'}^{(m)} \in \mathbb{R}^G$  corresponds to gene clusters that are only present in the $m$th study. In the next section, we extend the MSFA model to the nonlinear case while retaining this interpretability from the shared and study-specific loadings matrices.  

\subsection{Multi-study sparse deep generative model}\label{sec:methods-msdgm}

To allow for nonlinear relationships between the data and the latent factors, 
we propose a sparse deep generative model (DGM) for multi-study data. To retain the interpretability of a linear model, we introduce a sparse masking parameter  which allows the observed feature  to depend on a subset of the latent factors. 

\begin{figure}[H]
    \includegraphics[width=\textwidth]{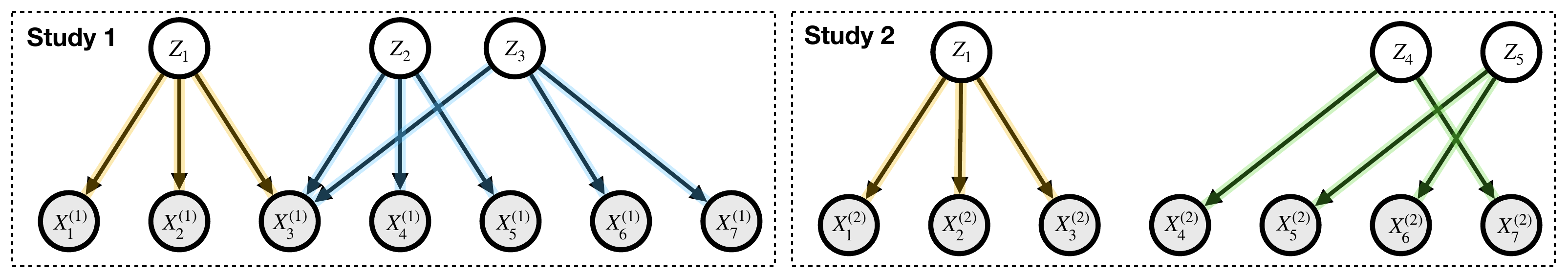}
    \caption{Example of a two-study sparse model: both study 1 and 2 have the same factor $z_1$ relationship, whereas factors $z_2,z_3$ are specific to study 1, and factors $z_4, z_5$ are specific to study 2. }\label{fig:sparse-example}
\end{figure}

More specifically, our sparse DGM for multi-study data is: for $i=1,\dots, n_m$ and $j=1,\dots, G$,
\begin{align}\label{eq:sparse-dgm-gauss}
x_{ij}^{(m)} = f_{\theta, j}(\widetilde{\bm{w}}_j^{(m)} \odot \widetilde{\bm{z}}_i^{(m)}) + \varepsilon_{ij}^{(m)}, \text{ where } \varepsilon_{ij}^{(m)} \stackrel{iid}{\sim} \mathcal{N}(0, \sigma_j^2), 
\end{align}
where $f_{\theta}:\mathbb{R}^{K}\to \mathbb{R}^G$ is a feedforward neural network parameterized by $\theta$. Above, we define $f_{\theta, j}(\bm{z})\defeq (f_{\theta}(\bm{z}))_j$ i.e., the $j$th output dimension of $f_{\theta}$. Further,  $\sigma_j^2$ is the noise variance for feature $j$, and $\odot$ denotes the element-wise product.  The sparse-per-feature vector $\widetilde{\bm{w}}_j^{(m)} \in\mathbb{R}^{K}$ selects which latent factors are used to produce feature $j$ in study $m$.  The vector $\widetilde{\bm{w}}_j^{(m)}$ combines shared and study-specific factor dimensions as:
\begin{align*}
\widetilde{\bm{w}}_j^{(m)} = 
\begin{pmatrix}
{\bm{w}}_j^{(S)} & \bm{0}_{1\times K_1}  & \cdots & \bm{0}_{1\times K_{m-1}}& {\bm{w}}_j^{(m)} & \bm{0}_{1\times K_{m+1}}&\cdots &\bm{0}_{1 \times K_M}
\end{pmatrix}^\top, 
\end{align*}
where $\bm{w}_{j}^{(S)}$ is shared across all studies, so that all studies select the same dimensions of the shared factor space to produce feature $j$. The vector ${\bm{w}}_j^{(m)}$ allows study $m$ to additionally use study-specific latent factor dimensions to produce feature $j$. See \Cref{fig:sparse-example} for an illustration. The zero padding in $\widetilde{\bm{w}}_j$ is included because we use the same $f_{\theta}$ for all studies and so require the input of $f_{\theta}$ to be partitioned into shared and study-specific dimensions. 

Similarly, the latent factors $\widetilde{\bm{z}}_i^{(m)}$ combine shared and study-specific dimensions: 
\begin{align*}
\widetilde{\bm{z}}_i^{(m)} =
\begin{pmatrix}
{\bm{z}}_{i}^{(m)} & \bm{0}_{1\times K_1}& \cdots & \bm{0}_{1\times K_{m-1}}& \bm{\zeta}_{i}^{(m)}& \bm{0}_{1\times K_{m+1}}&\cdots &\bm{0}_{1 \times K_M}
\end{pmatrix}^\top,
\end{align*}
where $\bm{z}_i^{(m)} \in \mathbb{R}^{K_S}$ are factors from the shared space and $\bm{\zeta}_i^{(m)}\in \mathbb{R}^{K_m}$ are factors from the study-specific space. Note that $\bm{z}_i^{(m)}$ are local variables specific to each sample $i$; these factors are ``shared'' in that modulating $\bm{z}_{i}^{(m)}$ for any $m$ will have the same effect on $\bm{x}_{i}^{(m)}$ (via the shared $\bm{w}^{(S)}$). 

For all studies $m=1,\dots,M$, the shared and study-specific factors are assigned standard normal priors:
\begin{align*}
\bm{z}_i^{(m)} \sim \mathcal{N}_{K_S}(\bm{0}, \bm{I}), \quad \bm{\zeta}_i^{(m)} \sim \mathcal{N}_{K_m}(\bm{0}, \bm{I}).
\end{align*}

The error variances $\sigma_j^2$ are assigned inverse-gamma priors:
\begin{align*}
\sigma_j^2 \sim \text{Inverse-Gamma}(\alpha,\beta).
\end{align*}

Similarly to \citet{moran2022identifiable}, the shared masking variables are assigned a sparsity-inducing prior, the spike-and-slab lasso \citep{rockova2018spike}. Specifically, for $j=1,\dots, G; k = 1,\dots, K_S$:
\begin{align*}
p(w_{jk}^{(S)}|\gamma_{jk}^{(S)}) &= \left(\frac{\lambda_1}{2} \exp(-\lambda_1 |w_{jk}^{(S)}|)\right)^{\gamma_{jk}^{(S)}}\left(\frac{\lambda_0}{2} \exp(-\lambda_0 |w_{jk}^{(S)}|)\right)^{(1-\gamma_{jk}^{(S)})}, \\ 
\gamma_{jk}^{(S)}|\eta_k^{(S)} &\sim \text{Bernoulli}(\eta^{(S)}_k), \\ 
\eta^{(S)}_k &\sim \text{Beta}(a_S, b_S).
\end{align*}
 That is, \emph{a priori} $w_{jk}^{(S)}$ is drawn from a Laplacian ``slab'' with parameter $\lambda_1$, or a Laplacian ``spike'' with parameter $\lambda_0$, where $\lambda_0 \gg \lambda_1$. The binary variable $\gamma_{jk}^{(S)}\in \{0, 1\}$ indicates whether $w_{jk}^{(S)}$ is drawn from the spike or slab; to allow for uncertainty over this draw, $\gamma_{jk}^{(S)}$ is assigned a Bernoulli distribution with parameter $\eta^{(S)}_k$.  The parameter $\eta^{(S)}_k$ reflects the proportion of $w_{jk}^{(S)}$ drawn from the slab, and is itself assigned a Beta prior with hyperparameters $a_S, b_S \in \mathbb{R}^+$ controlling the sparsity of $\gamma_{jk}^{(S)}$; this prior helps to ``zero out'' extraneous factor dimensions and consequently estimate $K_S$. Setting $a_S\propto 1/K$, $b_S = 1$ recovers the finite Indian Buffet Process prior \citep{griffiths2005infinite}.

For the study-specific masking variables $w_{jk}^{(m)}$, we similarly assign spike-and-slab lasso priors \citep{rockova2018spike} with associated hyperpriors for $\gamma_{jk}^{(m)}$, $\eta_k^{(m)}$ and hyperparameters $a_m, b_m$. Finally, we denote $\bm{W}^{(S)} = [\bm{w}_1^{(S)} \cdots \bm{w}^{(S)}_G]^T$, $\bm{\Gamma}^{(S)} = (\gamma_{jk}^{(S)})_{1\leq j\leq G, 1\leq k\leq K}$ (similarly for $\bm{W}^{(m)}$ and $\bm{\Gamma}^{(m)}$).

\subsection{Estimation}

We use approximate maximum \emph{a posteriori} (MAP) estimation for the parameters $\left\{\theta, \bm{W}^{(S)},  \bm{\eta}^{(S)}, \{ \bm{W}^{(m)}, \bm{\eta}^{(m)}\}_{m=1}^M, \bm{\Sigma}\right\}$ and amortized variational inference for the factors $\{\bm{z}_{1:n_m}^{(m)}, \bm{\zeta}_{1:n_m}^{(m)}\}_{m=1}^M$. This procedure for approximating the posterior is called a multi-study sparse VAE (MSSVAE).

The exact MAP objective (up to constant terms) is:
\begin{align}
\log p_{\theta}&\left(\bm{W}^{(S)}, \bm{\eta}^{(S)}, \left\{ \bm{W}^{(m)}, \bm{\eta}^{(m)}\right\}_{m=1}^M, \bm{\Sigma} \bigg|\{\bm{X}^{(m)}\}_{m=1}^M\right) \notag\\ 
&= \sum_{m=1}^M\sum_{i=1}^{n_m} \log \left[\int p_{\theta}(\bm{x}_i^{(m)}|\bm{W}^{(S)},\bm{W}^{(m)}, \bm{z}_i^{(m)}, \bm{\zeta}_i^{(m)},\bm{\Sigma}) p(\bm{z}_i^{(m)}) p(\bm{\zeta}_i^{(m)}) d\bm{z}_i^{(m)}d\bm{\zeta}_i^{(m)}\right]  \notag\\ 
&\quad\quad + \log p(\bm{\Sigma}) + \log \left[\int p(\bm{W}^{(S)}|\bm{\Gamma}^{(S)})p(\bm{\Gamma}^{(S)}|\bm{\eta}^{(S)})p(\bm{\eta}^{(S)}) d\bm{\Gamma}^{(S)} \right] \notag\\ 
&\quad\quad + \sum_{m=1}^M \log \left[\int p(\bm{W}^{(m)}|\bm{\Gamma}^{(m)})p(\bm{\Gamma}^{(m)}|\bm{\eta}^{(m)})p(\bm{\eta}^{(m)}) d\bm{\Gamma}^{(m)}\right]. \label{eq:exact_map}
\end{align}

The exact MAP objective is intractable; we lower bound the first term of \Cref{eq:exact_map} using variational inference. For the shared factors $\bm{z}_{i}^{(m)}$ and, separately for each study, the study-specific factors $\bm{\zeta}_i^{(m)}$, we use the variational families:
  \begin{align*}
q_{\psi_S}(\bm{z}_{i}^{(m)}|\bm{x}_i^{(m)}) &= N(\mu_{\psi_S}(\bm{x}_i^{(m)}), \mathrm{diag}(\sigma_{\psi_S}^2(\bm{x}_i^{(m)}))), \\
q_{\psi_m}(\bm{\zeta}_{i}^{(m)}|\bm{x}_i^{(m)}) &= N(\mu_{\psi_m}(\bm{x}_i^{(m)}), \text{diag}(\sigma_{\psi_m}^2(\bm{x}_i^{(m)}))),
  \end{align*}
  where $\mu_{\psi_S}, \sigma_{\psi_S}^{2}:\mathbb{R}^G\to\mathbb{R}^{K_S}$ and $\mu_{\psi_m}, \sigma_{\psi_m}^{2}:\mathbb{R}^G\to\mathbb{R}^{K_m}$ are neural networks parameterized by $\psi_S$ and $\psi_m$, respectively.

Next, we lower bound the last two terms of the MAP objective (\Cref{eq:exact_map}), again using variational inference:
\begin{align*}
q(\bm{\Gamma}^{(S)}) &= p(\bm{\Gamma}^{(S)}|\bm{W}_{\mathrm{old}}^{(S)}, \bm{\eta}_{\mathrm{old}}^{(S)}),\\ 
q(\bm{\Gamma}^{(m)}) &= p(\bm{\Gamma}^{(m)}|\bm{W}_{\mathrm{old}}^{(m)}, \bm{\eta}_{\mathrm{old}}^{(m)}), \quad m=1,\dots, M,
\end{align*}
where $\{\bm{W}_{\mathrm{old}}^{(\cdot)}, \bm{\eta}_{\mathrm{old}}^{(\cdot)}\}$ denote the updates from the previous iteration of the algorithm. This choice corresponds to the usual expectation in the EM algorithm. 

Bringing together these approximations, the MAP objective (\Cref{eq:exact_map}) has the lower bound:
\begin{align}
\mathcal{L}(\Theta) &= \sum_{m=1}^M\left\{\sum_{i=1}^{n_m} \left\{\mathbb{E}_{q_{\psi_S}(\bm{z}_{i}^{(m)}|\bm{x}_i^{(m)})q_{\psi_m}(\bm{\zeta}_{i}^{(m)}|\bm{x}_i^{(m)})}\left[ \log p_{\theta}(\bm{x}_i^{(m)} | \bm{W}^{(S)}, \bm{W}^{(m)}, \bm{z}_i^{(m)}, \bm{\zeta}_i^{(m)}, \bm{\Sigma}) \right] \right.\right. \notag\\
&\quad \left. -  D_{KL}(q_{\psi_S}(\bm{z}_{i}^{(m)}|\bm{x}_i^{(m)}) || p(\bm{z}_{i}^{(m)})) - D_{KL}(q_{\psi_m}(\bm{\zeta}_{i}^{(m)}|\bm{x}_i^{(m)}) || p(\bm{\zeta}_{i}^{(m)})) \right\}\notag\\
&\quad  \left. + \mathbb{E}_{\bm{\Gamma}^{(m)}|\bm{W}^{(m)}, \bm{\eta}^{(m)}}{[\log p(\bm{W}^{(m)}|\bm{\Gamma}^{(m)}) p(\bm{\Gamma}^{(m)}|\bm{\eta}^{(m)})p(\bm{\eta}^{(m)})]}\right\}
  \notag \\
&\quad   +  \mathbb{E}_{\bm{\Gamma}^{(S)}|\bm{W}^{(S)}, \bm{\eta}^{(S)}}{[\log p(\bm{W}^{(S)}|\bm{\Gamma}^{(S)}) p(\bm{\Gamma}^{(S)}|\bm{\eta}^{(S)})p(\bm{\eta}^{(S)})]} + \log p(\bm{\Sigma}) , \label{eq:ELBO-gauss}
\end{align}
where $\Theta = \{\theta, \psi_S, \psi_m, \bm{W}^{(S)}, \bm{W}^{(m)},\bm{\eta}^{(S)}, \bm{\eta}^{(m)}, \bm{\Sigma}\}_{m=1}^M$ and $D_{KL}$ is the Kullback-Leibler divergence.

To optimize \Cref{eq:ELBO-gauss}, we alternate between an expectation step and a maximization step. In the expectation step, we calculate expectations with respect to latent indicators $\{\bm{\Gamma}^{(S)}, \bm{\Gamma}^{(m)}\}_{m=1}^M$ and latent factors $\{\bm{z}_i^{(m)}, \bm{\zeta}_i^{(m)}\}_{i,m=1}^{n_m, M}$. In the maximization step, we take gradient steps in model parameters and variational parameters with the help of reparameterization gradients \citep{kingma2013auto,rezende2014stochastic}. For more details see the supplement (\Cref{sec:comp-det}).

\textbf{Choice of $K$.}  In practice, the true number of latent factors $K=K_S+K_1+\cdots+K_M$ is unknown, so we initialize with an overestimate of $K$; the sparsity-inducing prior on $\widetilde{\bm{W}}$ then encourages uninformative columns to be set to all zeros, and after removing these, the remaining column dimension is our estimate of $K$. This strategy is common in variational inference for Bayesian nonparametric models \citep{doshi2009variational}.

\subsection{Negative-Binomial MSSVAE}

To accommodate overdispersed count data as in the RNA-seq data example, we additionally consider a multi-study DGM with a negative-binomial likelihood. Specifically, the observed data is distributed as:
\begin{align}
\bm{x}_{ij}^{(m)}|\widetilde{\bm{w}}_j^{(m)},\widetilde{\bm{z}}_i^{(m)},\ell_i^{(m)},\phi_j \sim \text{NB}(\text{mean}=\ell_i^{(m)} f_{\theta, j}(\widetilde{\bm{w}}_j^{(m)} \odot \widetilde{\bm{z}}_i^{(m)}), \text{inv-disp} =\phi_j)  \label{eq:nb-mssvae}
\end{align}
where $\ell_i^{(m)}$ is the library size (i.e. sum of transcripts in sample $i$ of study $m$) and $\phi_j$ is the per-gene inverse dispersion. To ensure the mean is positive, the neural network $f_{\theta}$ applies to the final layer an element-wise softplus function: $\text{softplus}(z)=\log(1 + \exp(z))$.
We use a parameterization of the negative-binomial distribution with the following probability mass function (for mean $\mu$, inverse-dispersion $\phi$):
\begin{align*}
p_{NB}(x;\mu,\phi) = \frac{\Gamma(x+\phi)}{\Gamma(x+1)\Gamma(\phi)}\left(\frac{\phi}{\phi+\mu}\right)^{\phi}\left(\frac{\mu}{\phi+\mu}\right)^x.
\end{align*}
We choose this parameterization as it was noted to be computationally stable for genomics data \citep{lopez2018deep}. Also following \citet{lopez2018deep}, we take the prior on $\ell_i$ to be $\ell_i \sim \text{LogNormal}(\ell_\mu, \ell_\sigma^2)$, where $\ell_\mu, \ell_\sigma^2$ are set to the empirical mean and variance of the observed library sizes, $\{\sum_{j=1}^G x_{ij}\}_{i=1}^n$; we estimate $\ell_i$ using variational inference with variational distribution $q(\ell_i|\bm{x}_i) = \text{LogNormal}(\mu_{\ell}(\bm{x}_i), \sigma_{\ell}^2(\bm{x}_i))$, where $\mu_{\ell}, \sigma_{\ell}^2:\mathbb{R}^G\to\mathbb{R}$ are neural networks.

The negative-binomial MSSVAE is fit using the same objective as in \Cref{eq:ELBO-gauss}, replacing the Gaussian likelihood with the negative binomial, subtracting the Kullback-Leibler divergence $D_{KL}(q(\ell_i^{(m)}|\bm{x}_i^{(m)}) || p(\ell_i^{(m)}))$, and omitting the noise variance $\bm{\Sigma}$ terms. The inverse-dispersion parameters $\{\phi_j\}_{j=1}^G$ are not assigned a prior but are also estimated in the optimization of the MAP objective.

\section{Identifiability}\label{sec:id-theory}

Identifiability is important for reliability and interpretability of the latent variables. We first consider a single-study deep generative model with additive Gaussian noise and provide a new identifiability proof, strengthening the existing result from \citet{moran2022identifiable}. We next extend the results to the negative-binomial case and finally to the multi-study model, where we prove identifiability of the shared and study-specific masking matrices.

\subsection{Single-Study Model}

\subsubsection{Gaussian case}

A single-study generative model with additive Gaussian noise is:
\begin{align}
\bm{x}_i = f(\bm{z}_i) + \bm{\varepsilon}_i, \quad \bm{\varepsilon}_i\stackrel{ind}{\sim} N_G(\bm{0}, \bm{\Sigma}), \quad \bm{\Sigma}=\mathrm{diag}\{\sigma_j^2\}_{j=1}^G, \quad i =1,\dots, N, \label{eq:general-dgm}
\end{align}
 where $\bm{z}_i\sim \zpr \in \mathcal{P}_Z$ and  $f\in\mathcal{F}$, with $\mathcal{P}_z$ denoting a class of probability distributions and $\mathcal{F}$ denoting a class of functions.

There are three main identifiability issues in the model \Cref{eq:general-dgm}. First, the noise variance $\bm{\Sigma}$ is unknown. Second, the  latent factor dimension is generally unknown. Third, we can always re-parameterize the model:
\begin{align}
\bm{x}_i = f(h^{-1}(h(\bm{z}_i))) + \bm{\varepsilon}_i, \quad \text{for any invertible }h: \mathbb{R}^K \to\mathbb{R}^K,  \label{eq:single-indeterminancies}
\end{align}
when $f\circ h^{-1}\in \mathcal{F}$ and $h_{\#}\zpr\in\mathcal{P}_Z$, where $h_{\#}\zpr$ is the pushforward measure of $\zpr$ by $h$. To restrict possible functions $h$, we can place restrictions on $\mathcal{F}$, $\mathcal{P}_Z$, or both \citep{xi2023indeterminacy,moran2026towards}.    

Our strategy is to restrict $\mathcal{F}$ to a specific set of sparse functions. As discussed in \Cref{sec:methods-msdgm}, we  do so by introducing the parameter $\bm{w}_j$ which selects which of the $K$ latent factor dimensions are used to generate $x_{ij}$:
\begin{align}
{x}_{ij} = f_j(\bm{w}_j\odot\bm{z}_i) + {\varepsilon}_{ij}, \quad i=1,\dots, N;\  j=1,\dots, G, \label{eq:single-study-dgm}
\end{align}
where $\bm{w}_j\in \mathbb{R}^K$. If $w_{jk}=0$, then $z_{ik}$ cannot contribute to $x_{ij}$. We denote $\bm{W}=[\bm{w}_1,\dots, \bm{w}_G]^T$. 

We assume a specific kind of sparsity in $\bm{W}$, corresponding to an anchor feature assumption. An anchor feature is a coordinate of $\bm{x}_i$ that depends only on one latent factor dimension. Anchor feature assumptions have also been used for identification in latent topic modeling \citep{arora2013practical}, linear factor analysis \citep{bing2020adaptive, bing2023detecting} and nonlinear factor analysis \citep{moran2022identifiable,xu2023identifiable}.

\begin{assumption}[Anchor features.]
For every non-zero factor dimension $\bm{{z}}_{\cdot k}$, there are at least two features $\bm{x}_{\cdot j(k)}$, $\bm{x}_{\cdot j'(k)}$ which depend only on that factor. Moreover, the two features have the same mapping from the factors, up to a multiplicative factor $c_{j'(k)}\in\mathbb{R}$; that is, for all $i=1,\dots, N$:
\begin{align}
\mathbb{E}[x_{i,j(k)}|\bm{{z}}_i] = f_{j(k)}({z}_{ik}),\qquad \mathbb{E}[x_{i,j'(k)}|{\bm{z}}_{i}] = c_{j'(k)} f_{j(k)}({z}_{ik}),
\end{align}
where $f_{j(k)}:\mathbb{R}\to\mathbb{R}$ are strictly monotone functions.
We refer to such features as ``anchor features''. Note we have slightly abused notation by using $f_{j(k)}({z}_{ik})$ to denote $f([0, \dots, 0, {w}_{jk}, 0,\dots, 0] \odot [0,\dots, 0,{z}_{ik},0,\dots, 0])_j$; that is, $\bm{w}_j$ has only one non-zero entry in dimension $k$.  
\label{ass:anchor}
\end{assumption}

\begin{remark}
Assumption \ref{ass:anchor} differs from \citet{moran2022identifiable} in that anchor feature means need only be equal up to a multiplicative constant. However, we require
strictly monotone anchor mappings $f_{j(k)}$, which we use to identify the support of $\bm{W}$.
\end{remark}

Our proof strategy extends the ``parallel rows'' argument of \citet{bing2023detecting}, who show for a linear factor model that rows of the marginal population correlation matrix (excluding diagonals) are parallel if and only if the corresponding rows of the loadings matrix are parallel (assuming the latent factor covariance is rank $K$). \Cref{ass:non-parallel} extends this to the nonlinear setting by assuming non-anchor features are non-parallel almost everywhere, and \Cref{ass:cov} is the analogue of rank-$K$ covariance; together, these ensure that parallel rows of the correlation matrix must correspond to anchor features, allowing anchors to be identified from the marginal correlation matrix.

\begin{assumption}[Non-anchor feature differences] \label{ass:non-parallel}
$\forall l,p$, $\forall c\neq 0$, if
\[
f_l(\bm{w}_l\odot \bm{z}_i)=cf_p(\bm{w}_p\odot \bm{z}_i) \text{ a.e. }
\]
then $l,p$ are anchors for the same factor.
\end{assumption}

\begin{assumption}[Covariance non-degeneracy] \label{ass:cov}
For every \(l\neq p\) and \(c\neq 0\), if
\[
\operatorname{Cov}\{f_l(\bm{w}_l\odot \bm{z}_i)-c f_p(\bm{w}_p\odot \bm{z}_i),\bm{x}_{ir}\}=0
\quad\text{for all }r\notin\{l,p\},
\]
then
\[
f_l(\bm{w}_l\odot \bm{z}_i)=c f_p(\bm{w}_p\odot \bm{z}_i)\quad\text{a.e.}
\]
Further, for every anchor feature $x_{i,j(k)}$, there is at least one other feature $x_{ir}$ (in addition to $x_{i,j'(k)}$) such that $\text{Cov}(x_{ij(k)},x_{ir})\neq 0$.
\end{assumption}

\begin{assumption}[Conditional latent non-degeneracy]\label{ass:cond-non-degen}
The conditional distribution $p(z_{ik}|\bm{z}_{i, \backslash k})$ is non-atomic almost everywhere, where $\bm{z}_{i,\backslash k} = (z_{i1},\dots, z_{i,k-1}, z_{i,k+1}, \dots, z_{iK})$.
\end{assumption}

\begin{assumption}[Functional non-degeneracy]\label{ass:fun-non-degen}
For any $j\in\{1,\dots, G\}$ and $k\in\{1,\dots, K\}$, if $w_{jk}\neq 0$, then 
\[
    P\left[
        \operatorname{Var}\left\{
            f_j(w_j\odot z_i)\mid \bm{z}_{i,\setminus k}
        \right\}>0
    \right]>0.
\]
\end{assumption}

\Cref{ass:cond-non-degen,ass:fun-non-degen} are akin to a faithfulness assumption, ensuring that if $w_{jk}\neq 0$, then $x_{ij}$ depends non-trivially on $z_{ik}$.

\begin{theorem}\label{thm:anchor-id}
Consider the model in \Cref{eq:general-dgm}. Suppose \Cref{ass:anchor,ass:non-parallel,ass:cov,ass:cond-non-degen,ass:fun-non-degen} hold. Then
\begin{enumerate}
    \item The anchor features and the latent dimensionality $K$ are identifiable from the marginal population correlation matrix.
    \item For any two solutions $(f, \bm{W}, \bm{\Sigma}, p_z)$, $(\widehat{f}, \widehat{\bm{W}}, \widehat{\bm{\Sigma}}, p_{\widehat{z}})$ with
    \begin{align*}
p(\bm{x}_i | f, \bm{W},\bm{\Sigma}) = p(\bm{x}_i|\widehat{f}, \widehat{\bm{W}},\widehat{\bm{\Sigma}}),
    \end{align*}
    the factors are equal in distribution up to element-wise transformations and permutations:
    \begin{align*}
(z_{i1},\dots, z_{iK}) \stackrel{d}{=} (h_1(\widehat{z}_{i,\pi(1)}),\dots, h_K(\widehat{z}_{i,\pi(K)})),
\end{align*}
where $h_k:\mathbb{R}\to\mathbb{R}$ are invertible functions and  $\pi:\{1,\dots, K\}\to\{1,\dots, K\}$ is a permutation.
    \item The support of $\bm{W}$ is identifiable, up to permutation of columns.
    \item The error variances $\bm{\Sigma}$ are identifiable.
\end{enumerate}
\end{theorem}

The proof is in the supplement (\Cref{sec:proofs}); we provide a proof sketch here. (1) Parallel rows of the marginal correlation matrix (excluding diagonals) correspond to anchor features, identifying the anchor features as rows of $\bm{W}$ with one non-zero element. (2) Since anchor features are identified and parallel, we identify the noise variances corresponding to them. (3) We then prove the factors are equal in distribution up to an element-wise transform and permutation. (4) By contradiction, under \Cref{ass:cond-non-degen,ass:fun-non-degen}, the support of $\bm{W}$ is identifiable. (5) Finally, the non-anchor noise variances are identified.

\begin{remark}
Our anchor feature identification strategy follows the constructive proof of \citet{bing2023detecting}; in practice, rather than selecting anchors from the marginal correlation matrix, we place a sparsity-inducing prior on $\bm{W}$, which empirically outperforms fixing the anchor rows directly. We anticipate this is due to ``softer'' regularization approaches performing better than hard restrictions for neural network optimization \citep{wilson2025position}. Further, the proof requires no assumptions on $p_z$ beyond the non-degeneracy conditions in \Cref{ass:cov,ass:cond-non-degen}; for estimation we use standard Gaussian priors, which perform as well as learned-covariance priors at lower cost.
\end{remark}

Our identifiability results hold for deep generative models with additive Gaussian noise. We anticipate that these arguments can be extended to other additive noise models. 

\subsubsection{Negative-binomial case}

We now consider the single-study negative-binomial model with unknown dispersions $\{\phi_j\}_{j=1}^G$ and known and fixed library sizes $\{\ell_i\}_{i=1}^N$:
\begin{align}
{x}_{ij}|{\bm{w}}_j,{\bm{z}}_i,\ell_i,\phi_j \sim \text{NB}(\text{mean}=\ell_i f_{j}({\bm{w}}_j \odot {\bm{z}}_i), \text{inv-disp} =\phi_j).  \label{eq:nb-svae}
\end{align}
Denote $\bm{\phi} = (\phi_1,\dots, \phi_G)$.

For the negative-binomial model above, we adapt the anchor feature assumption slightly to account for the library size.

\begin{assumption}[Negative-binomial anchor features.]
For every non-zero factor dimension $\bm{{z}}_{\cdot k}$, there are at least two features $\bm{x}_{\cdot j(k)}$, $\bm{x}_{\cdot j'(k)}$ which depend only on that factor. Moreover, the two features have the same mapping from the factors, up to a multiplicative factor $c_{j'(k)}> 0$; that is, for all $i=1,\dots, N$:
\begin{align}
\mathbb{E}[x_{i,j(k)}|\bm{{z}}_i] = \ell_if_{j(k)}({z}_{ik}),\qquad \mathbb{E}[x_{i,j'(k)}|{\bm{z}}_{i}] = c_{j'(k)} \ell_i f_{j(k)}({z}_{ik}),
\end{align}
where $f_{j(k)}:\mathbb{R}\to\mathbb{R}^+$ are strictly monotone functions.
\label{ass:nb-anchor}
\end{assumption}

\begin{theorem}\label{thm:nb-id}
Consider the negative binomial model \Cref{eq:nb-svae}. Suppose that the library sizes $\{\ell_i\}_{i=1}^N$ are known and fixed. Suppose \Cref{ass:non-parallel,ass:cov,ass:cond-non-degen,ass:fun-non-degen,ass:nb-anchor} hold. Then
\begin{enumerate}
    \item The anchor features and the latent dimensionality $K$ are identifiable from the marginal population correlation matrix.
    \item For any two solutions $(f, \bm{W}, p_z, \bm{\phi})$, $(\widehat{f}, \widehat{\bm{W}}, p_{\widehat{z}}, \widehat{\bm{\phi}})$ with
    \begin{align*}
p(\bm{x}_i | f, \bm{W}, \phi, \ell_i) = p(\bm{x}_i|\widehat{f}, \widehat{\bm{W}}, \widehat{\phi}, \ell_i),
    \end{align*}
    the factors are equal in distribution up to element-wise transformations and permutations:
    \begin{align*}
(z_{i1},\dots, z_{iK}) \stackrel{d}{=} (h_1(\widehat{z}_{i,\pi(1)}),\dots, h_K(\widehat{z}_{i,\pi(K)})),
\end{align*}
where $h_k:\mathbb{R}\to\mathbb{R}$ are invertible functions and  $\pi:\{1,\dots, K\}\to\{1,\dots, K\}$ is a permutation.
    \item The support of $\bm{W}$ is identifiable, up to permutation of columns.
    \item The inverse-dispersion parameters are identifiable. 
\end{enumerate}
\end{theorem}

The proof is in the supplement (\Cref{sec:proofs}). The proof strategy is similar to \Cref{thm:anchor-id}. The main difference is that we cannot use noise deconvolution arguments to identify the distribution of the latent factors; instead, we rely on identifiability of mixtures of power-series distributions \citep{sapatinas1995identifiability}.

\begin{remark}
\Cref{thm:nb-id} assumes known, fixed library sizes, whereas the NB-MSSVAE treats them as latent. We leave identifiability under latent library sizes to future work.
\end{remark}

\subsection{Multi-Study Model}

We return to the multi-study sparse deep generative model:
\begin{align}
x_{ij}^{(m)} = f_j(\widetilde{\bm{w}}_j^{(m)}\odot \widetilde{\bm{z}}_i^{(m)}) + \varepsilon_{ij}^{(m)}, \quad \bm{\varepsilon}_{i}^{(m)} \stackrel{ind}{\sim} N_G(\bm{0}, \bm{\Sigma}),\label{eq:thm-model}
\end{align}
where again $\bm{\Sigma}=\mathrm{diag}(\sigma_1^2,\dots,\sigma_G^2)$, $\widetilde{\bm{W}}^{(m)}$ concatenates the shared matrix $\bm{W}^{(S)}$ with the study-specific matrix $\bm{W}^{(m)}$, and $\widetilde{\bm{z}}_i^{(m)}$ concatenates the shared factors $\bm{z}_i^{(m)}\sim \zpr\in\mathcal{P}$ with the study-specific factors $\bm{\zeta}_i^{(m)}\sim p_{\zeta^{(m)}}\in\mathcal{P}$, so that shared factors are drawn from the same $\zpr$ across studies while study-specific factors may follow different distributions $p_{\zeta^{(m)}}$.

The multi-study model introduces an identifiability concern beyond the single-study case: the model in \Cref{eq:thm-model} is observationally equivalent to a misspecified model with no shared components ($\widehat{\bm{W}}^{(S)}=\bm{0}$), where the shared signal is instead concatenated with each study-specific matrix — an issue \citet{chandra2025inferring} call ``information switching.''

For the multi-study deep generative model, we prove that we can identify the shared signal from the study-specific signal under a canonical definition of shared and study-specific signal.

\begin{definition}[Canonical shared support decomposition]
For each study \(m\), let
\[
\mathcal C_m
=
\left\{
\operatorname{supp}(\widetilde{\bm W}^{(m)}_{\cdot k})
:
k=1,\dots,K_S+K_m
\right\}
\]
denote the set of column supports of
\(\widetilde{\bm W}^{(m)}=[\bm W^{(S)},\bm W^{(m)}]\), after removing
zero columns. Define the canonical shared support set, $\mathcal{C}_S^\mathrm{can}$, and study-specific support set for study $m$, $\mathcal{C}_m^\mathrm{can}$, by
\[
\mathcal C_S^{\mathrm{can}}
=
\bigcap_{m=1}^M \mathcal C_m, \quad \mathcal C_m^{\mathrm{can}}
=
\mathcal C_m \setminus \mathcal C_S^{\mathrm{can}}.
\]
\end{definition}

\begin{theorem}\label{thm:shared-id}
Suppose \Cref{ass:anchor,ass:non-parallel,ass:cov,ass:cond-non-degen,ass:fun-non-degen} hold for each study, considered separately. Then, the canonical supports of the shared matrix $\bm{W}^{(S)}$ and study-specific matrices $\bm{W}^{(m)}$ are identifiable. 
\end{theorem}

The proof is in the supplement (\Cref{sec:proofs}). First, we consider each study separately and utilize \Cref{thm:anchor-id} to obtain identifiability of the supports of each $\widetilde{\bm{W}}^{(m)}$, up to permutation of columns. (Note that this does not yet distinguish the shared from the study-specific signal due to column permutations.) Second,  we construct the shared matrix $\bm{W}^{(S)}$ from the columns that are equal across all studies. The remaining columns form the study-specific matrices $\{{\bm{W}}^{(m)}\}_{m=1}^M$. This proof is constructive and similar in spirit to \citet{sturma2023unpaired, mauri2025spectral}.

\begin{remark}
\citet{chandra2025inferring} show that, in the linear setting, information switching occurs if and only if the study-specific column spaces intersect non-trivially or $\mathrm{rank}(\bm{W}^{(m)}) < K_m$ for all $m$; our result is consistent with this, since the anchor assumption gives each $\bm{W}^{(m)}$ full rank.
\end{remark}

\begin{remark}
Although the proof is constructive, in practice we induce sparsity in $\bm{W}^{(S)}, \{\bm{W}^{(m)}\}_{m=1}^M$ via a Bayesian prior, which favors placing shared signal in $\bm{W}^{(S)}$: representing it instead across all $\{\bm{W}^{(m)}\}_{m=1}^M$ requires $M-1$ extra non-sparse columns.
\end{remark}

\section{Simulation studies} \label{sec:sim-study}

We consider two synthetic data experiments: nonlinear factor analysis with Gaussian observations (\Cref{sec:gaussian-sim}) and  synthetic bulk RNA-sequencing data with gene correlation (\Cref{sec:rna-sim}).  For the Gaussian case, we compare to VI-MSFA \citep{hansen2023fast}. For synthetic bulk RNA-seq data, we compare the NB-MSSVAE with multiGroupVI \citep{weinberger2022disentangling}. Note that VI-MSFA is not designed for count data, and multiGroupVI is not designed for Gaussian data.

\subsection{Gaussian nonlinear factor analysis}\label{sec:gaussian-sim}

\parhead{Synthetic data.} In this simulation study, the observed data is Gaussian, drawn from a nonlinear latent factor model. 
We have $M=3$ studies, $N_m=1000$ samples per study, $G=100$ features, $K_S=8$ shared latent dimensions and $K_m=2$ study-specific latent dimensions for each $m\in\{1,\dots, M\}$.

\begin{figure}[!ht]
\centering

\footnotesize\textbf{(a)}\par
\includegraphics[width=\textwidth]{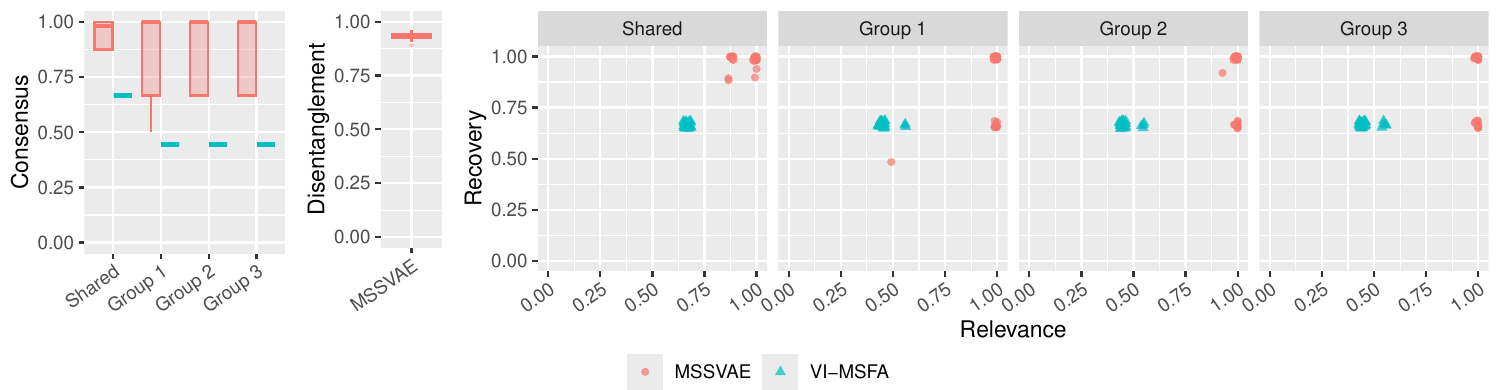}

\vspace{0.2cm}

\begin{minipage}[b]{0.49\textwidth}
\centering
\footnotesize\textbf{(b)}\par
\includegraphics[width=0.75\textwidth]{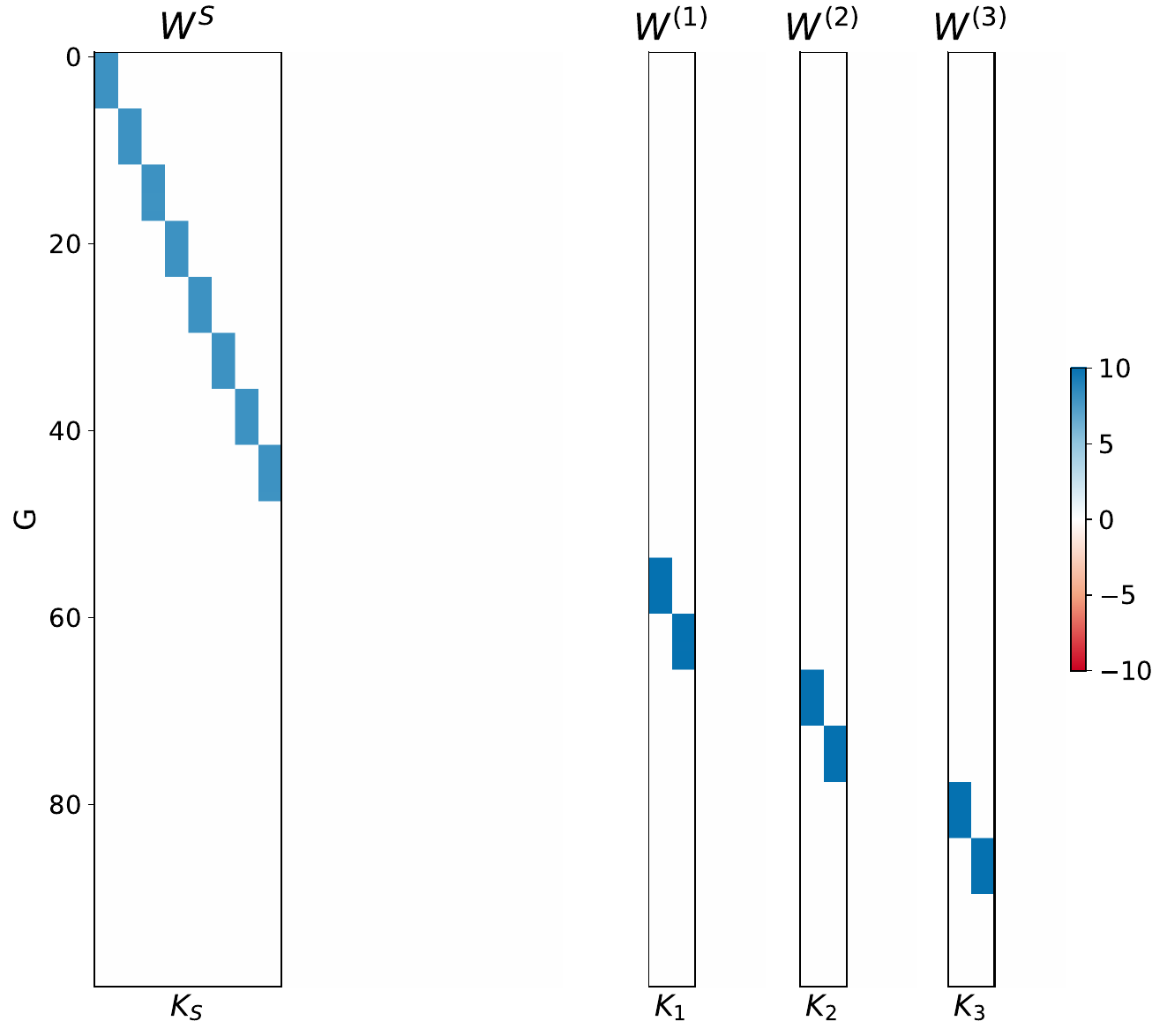}
\end{minipage}
\hfill
\begin{minipage}[b]{0.49\textwidth}
\centering
\footnotesize\textbf{(c)}\par
\includegraphics[width=0.75\textwidth]{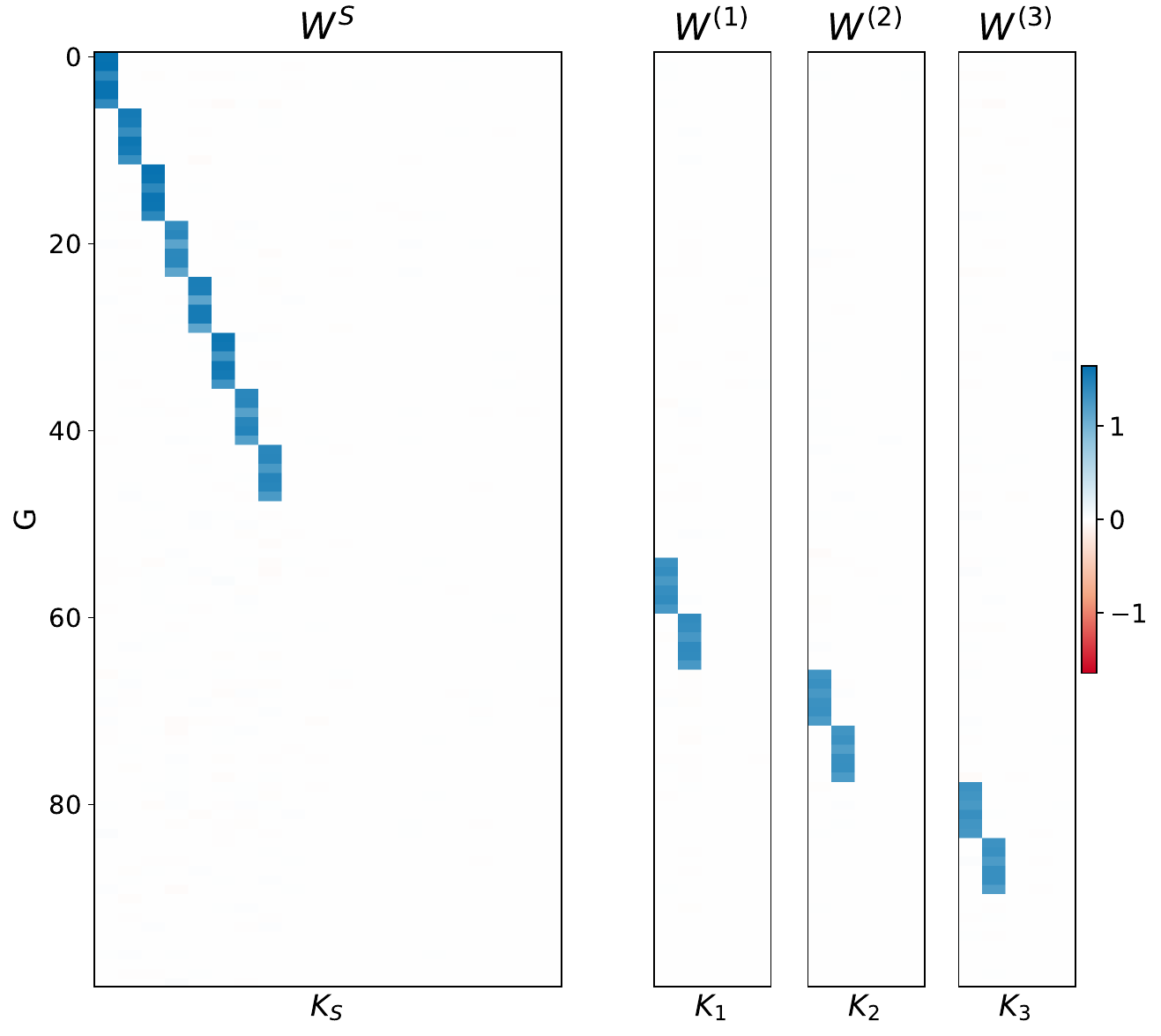}
\end{minipage}

\vspace{0.2cm}

\begin{minipage}[t]{0.49\textwidth}
\centering
\footnotesize\textbf{(d)}\par
\includegraphics[width=0.99\textwidth]{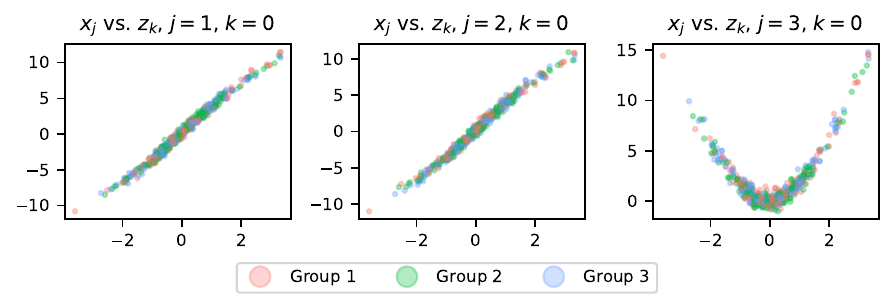}
\end{minipage}
\hfill
\begin{minipage}[t]{0.49\textwidth}
\centering
\footnotesize\textbf{(e)}\par
\includegraphics[width=0.99\textwidth]{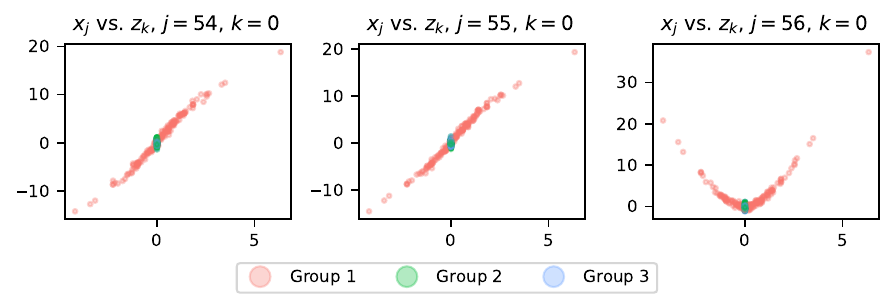}
\end{minipage}

\caption{Gaussian nonlinear factor analysis. (a) MSSVAE results over 30 different synthetic datasets. Consensus and relevance/recovery are separated into shared and group structure; disentanglement is over all factors. All metrics are in $[0,1]$; higher is better. (b) True $\bm{W}^S, \bm{W}^{(m)}$. (c) Estimated $\widehat{\bm{W}}^{(S)}, \widehat{\bm{W}}^{(m)}$. (d) Data vs. shared factors. (e) Data vs. study-1 factors. MSSVAE successfully estimates $\bm{W}$, including factor dimension (after removing columns of all zeros).}
\label{fig:gaussian_nonlinear}
\end{figure}

In this data, every third feature has a nonlinear relationship with the latent factors. Specifically, for  $m=1,\dots, M$,  $i=1,\dots, N_m$ and  $j=1,\dots, G$:
    \begin{flalign*}
x_{ij}^{(m)} &= 
\begin{cases}
\sum_{k=1}^{K_S} W_{jk}^{(S)} z_{ik}^{(m)} + \sum_{k=1}^{K_m} W_{jk}^{(m)} \zeta_{ik}^{(m)} +\varepsilon_{ij}^{(m)} &\text{if } j \not\equiv 0 \mod 3\\
\sum_{k=1}^{K_S} W_{jk}^{(S)} (z_{ik}^{(m)})^2 + \sum_{k=1}^{K_m} W_{jk}^{(m)} (\zeta_{ik}^{(m)})^2 +   \varepsilon_{ij}^{(m)} &\text{if } j \equiv 0 \mod 3
\end{cases}
    \end{flalign*}
where $\varepsilon_{ij}^{(m)} \sim \mathcal{N}(0, 0.2)$. The latent factors are generated as $\bm{z}_i^{(m)} \sim \mathcal{N}_{K_S}(\bm{0}, 0.5^2 \cdot \bm{I})$ and $\bm{\zeta}_i^{(m)} \sim \mathcal{N}_{K_m}(\bm{0}, 0.5^2 \cdot \bm{I})$. The matrices $\bm{W}^{(S)}, \{\bm{W}^{(m)}\}_{m=1}^M$ have a sparse structure displayed in \Cref{fig:gaussian_nonlinear}(b). The non-zero entries are ${w}_{jk}^{(S)} \stackrel{iid}{\sim} N(8, 0.01)$ and ${w}_{jk}^{(m)}\stackrel{iid}{\sim}N(10,0.01)$. 

\parhead{Metrics.} To measure performance of the MSSVAE, we report the consensus score \citep{H10}, relevance and recovery \citep{P06}, and the disentanglement score \citep{eastwood2018framework}. The consensus score measures how similar the estimated $\widehat{\bm{W}}$ are to the support of the true $\bm{W}$, penalizing overestimation of the support. Relevance measures how similar on average the estimated column supports are to the true support (cf. precision). Recovery instead measures how similar on average the true column supports are to the estimated supports (cf. recall). The disentanglement score is an average measure of how relevant each estimated factor is for a true factor, penalizing estimated factors that are informative of multiple true factors (i.e. ``entangling'' multiple true factors). For precise definitions, see the supplement (\Cref{sec:emp-det}). 
We compare MSSVAE to VI-MSFA \citep{hansen2023fast}. VI-MSFA uses variational inference to fit the linear multi-study factor analysis model \Cref{eq:unpaired-linear}.   VI-MSFA does not output a sparse loadings matrix. To obtain consensus, relevance and recovery scores for VI-MSFA, we apply a varimax rotation to the estimated loadings matrix  \citep{kaiser1958varimax} and retain values above a threshold. 

\parhead{Results.} The MSSVAE has high consensus, relevance and recovery, and disentanglement scores, measured on 30 different simulated datasets (\Cref{fig:gaussian_nonlinear}(a)). We visualize the output of one of the 30 experiments in \Cref{fig:gaussian_nonlinear}(c); here, MSSVAE successfully estimates the sparse structure and nonlinear mapping, even when using an overestimate of the true latent dimensions. In the supplement, we display the worst performing of the 30 experiments in terms of average consensus score (\Cref{sec:add-gauss-result}). 

VI-MSFA consistently has worse consensus, relevance and recovery scores relative to MSSVAE (\Cref{fig:gaussian_nonlinear}(a)); this is expected as VI-MSFA is a linear method, and cannot capture the nonlinear structure in this problem.

\subsection{Synthetic Bulk RNA-Sequencing Data}\label{sec:rna-sim}

In this simulation study, we consider synthetic bulk RNA-sequencing data.  This synthetic data is generated similarly to \citet{zappia2017splatter}, omitting dropout variation, which is specific to single-cell data. We compare the NB-MSSVAE to multiGroupVI \citep{weinberger2022disentangling}.

\parhead{Synthetic data.} We have $N=3200$ samples, $G=5000$ features and $M=3$ groups, with the proportion of samples in each group set to (30\%, 30\%, 40\%).

The base expression for each gene is a nonlinear function of the shared and study-specific latent factors (every third gene has a sine nonlinearity, analogous to \Cref{sec:gaussian-sim}), with loadings matrices $\bm{W}^{(S)}, \{\bm{W}^{(m)}\}_{m=1}^M$ following a mostly block-sparse structure (see \Cref{fig:syn-gene-W}(a) in the supplement for an example). This base expression is then normalized to sum to one within each sample; this, combined with a per-sample library size and a mean-variance trend typical of RNA-sequencing data \citep{mccarthy2012differential,zappia2017splatter}, determines the mean and dispersion of a negative-binomial draw for the observed counts. Full details are in the supplement (\Cref{sec:add-res}).

To this data, we fit the Negative-Binomial MSSVAE (\Cref{eq:nb-mssvae}). We initialize the latent dimensions to be $\widetilde{K}^{(S)}=50$ and $\widetilde{K}^{(m)}=10$ for all $m=1,\dots, M$. 

\parhead{Metrics.} For 30 different generated datasets, we fit the NB-MSSVAE and multiGroupVI. For NB-MSSVAE and multiGroupVI, we calculate the disentanglement scores and the area under the precision-recall curve (AUPRC). We compare AUPRC instead of consensus, relevance and recovery because multiGroupVI does not provide per-factor gene sets. Instead, multiGroupVI ranks genes by a likelihood-ratio score comparing models with/without shared or group-specific factors \citep{weinberger2022disentangling}. For NB-MSSVAE, genes were ranked by  the magnitude of each gene's shared or group-specific loading vector weighted by its posterior inclusion probability under the
  spike-and-slab prior, e.g. $\lVert \widehat{\bm{w}}_g^{(S)} \odot
  \bm{p}^{*(S)}_g\rVert^2$.  AUPRC was preferred over AUROC
  because the gene sets are highly imbalanced (around 600 shared genes out of 5,000 total).  For NB-MSSVAE, we additionally calculate the consensus, relevance and recovery  scores. Due to the correlation induced among genes by the normalization, there are some columns of $\bm{W}^{(S)}$ that are dense; consequently, we remove columns with greater than 25\% non-zero elements when calculating the consensus, relevance and recovery scores. 

\begin{figure}
\centering
\includegraphics[width=0.75\textwidth]{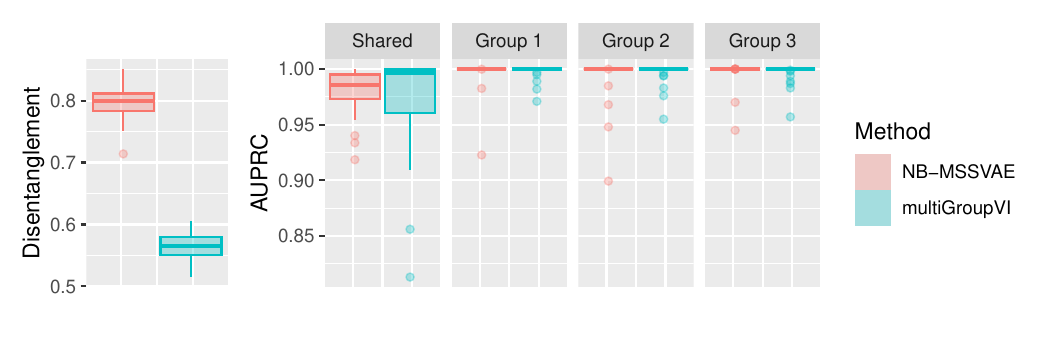}
\caption{NB-MSSVAE vs. multiGroupVI results over 30 different synthetic datasets. NB-MSSVAE has much higher disentanglement scores than multiGroupVI; the AUPRC values are comparable.}\label{fig:syn_gene_comparison}
\end{figure}

\parhead{Results.} NB-MSSVAE has much higher disentanglement scores than multiGroupVI (\Cref{fig:syn_gene_comparison}). This is because NB-MSSVAE can disentangle individual factor dimensions, whereas multiGroupVI can only separate into shared vs. group-specific blocks of factors. When we compare genes partitioned into shared vs. group-specific sets via the AUPRC, NB-MSSVAE and multiGroupVI perform similarly (\Cref{fig:syn_gene_comparison}). 

\begin{figure}
\centering
\includegraphics[width=\textwidth]{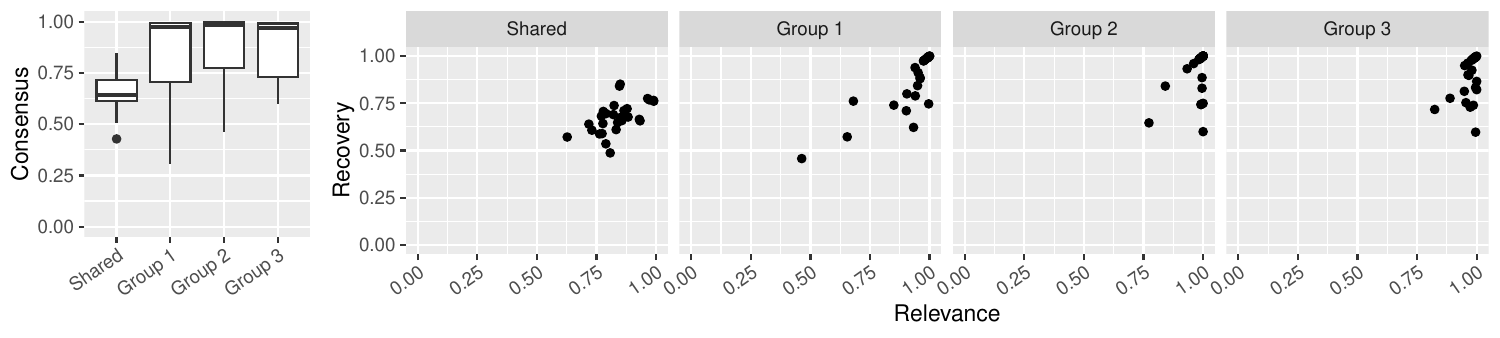}
\caption{Synthetic gene expression data: MSSVAE results over 30 different synthetic datasets. Consensus and relevance/recovery are separated into shared and group structure. All metrics are in $[0,1]$; higher is better.}
\label{fig:syn-gene-metrics}
\end{figure}

The NB-MSSVAE has good consensus, relevance and recovery scores over 30 different synthetic datasets, albeit lower than the Gaussian example (\Cref{fig:syn-gene-metrics}). This performance gap is due to the more difficult setting with overdispersed count data, per-sample library size effects, and correlation between genes induced by the per-sample normalization.  We visualize the output of one of the 30 experiments in the supplement (\Cref{fig:syn-gene-W}); here, the NB-MSSVAE estimates the sparse structure well, although the shared latent dimension is slightly overestimated.

\section{Platelet Data} \label{sec:platelet}

In this section, we analyze platelet transcriptomes from a multi-disease cohort of $N = 1,463$ ancestrally diverse patients. This dataset, curated from publications \citep{sol2020tumor,best2015rna,best2017swarm,best2018tumor,best2019rna}, spans $M = 6$ groups: cardiovascular disease, multiple sclerosis (immune-mediated), non-small-cell lung cancer, glioblastoma, other cancers (Table 1), contrasted with healthy donors. This disease spectrum enables systematic investigation of shared platelet molecular pathways versus disease-specific transcriptomic signatures across conditions ranging from cardiovascular dysfunction to malignancy-associated thrombosis. We consider the top $G = 5000$ most variable genes based on Pearson residuals of a negative binomial offset model \citep{lause2021analytic}, implemented using {\tt scanpy} \citep{wolf2018scanpy}. This filtering step enriches for genes with meaningful biological variation across disease states and individuals, prioritizing genes likely to capture disease-related and platelet-specific transcriptomic signatures while reducing noise from constitutively expressed genes.

\begin{table}[ht]
\centering
\footnotesize
\begin{tabular}{l r}
\hline
{\bf Group }& {\bf Count} \\
\hline
Healthy                   & 437 \\
Cardiovascular Disease    & 61  \\
Multiple Sclerosis         & 92  \\
Non Small Cell Lung Cancer     & 299 \\
Glioblastoma              & 286 \\
Other Cancer                    & 288 \\
\hline
Total & 1463 \\
\hline 
\end{tabular}
\caption{Platelet RNA-seq cohort composition by disease group (N=1,463).}
\label{tab:group_counts}
\end{table}

To this data, we fit the NB-MSSVAE and obtain $\widehat{\bm{W}}^{(S)}, \{\widehat{\bm{W}}^{(m)}\}_{m=1}^M$.  For each factor $k$, we select genes with $W_{gk}$ values above a threshold: $\mathcal{C}_k = \{ g \in \{1,\dots,G\}: |W_{g, k}| > 0.5\}$. 

To evaluate the biological significance of these gene sets $\{\mathcal{C}_k\}_{k=1}^K$, we conduct gene over-representation tests  \citep{xu_using_2024}. That is, we test whether known gene pathways occur in the selected set $\mathcal{C}_k$ more often than expected by chance using Fisher's exact test, with $p$-values adjusted for multiple testing using the Benjamini-Yekutieli procedure \citep{benjamini2001control}. For the known gene groups, we use Gene Ontology Biological Processes. The tests were implemented using {\tt clusterProfiler} \citep{xu_using_2024}.

Of the gene sets corresponding to the shared factor dimensions, 96\% were significantly enriched for at least one Gene Ontology Biological Process (FDR $<0.05$).  The enriched processes include fundamental platelet functions: hemostasis, thrombosis, inflammatory response, and immune response, along with essential cellular processes including cell metabolism, protein synthesis, and cell-cycle regulation. For each cluster, \Cref{fig:platelet-shared} in the supplement shows the top 3 biological processes  for each latent by effect size (i.e. fold enrichment), after filtering for processes appearing in more than 10\% of clusters. These shared factors represent conserved molecular pathways active across all disease states, likely reflecting core platelet biology and ubiquitous systemic stress responses.

Of the disease-specific clusters, 56\% were enriched for at least one Gene Ontology Biological Process (FDR $< 0.05$); the lower percentage is due to many clusters containing fewer than 5 genes (10/60). The biological processes are broadly consistent with the disease state: healthy clusters are enriched for housekeeping-type processes (e.g. RNA processing); cardiovascular clusters for oxidative stress (response to hydrogen peroxide); multiple sclerosis (immune-mediated) clusters for immune-related processes such as Type I interferon signaling, response to viruses, and innate immune response; and cancer clusters (glioblastoma, NSCLC, other cancer) for stress signaling (e.g. intrinsic apoptotic signaling) and inflammation (NF-kappaB signaling).
For each cluster, \Cref{fig:platelet-specific} in the supplement shows the top 5 biological processes by fold enrichment, after filtering for processes appearing in more than 10\% of clusters.  This separation of shared and disease-specific biological processes demonstrates the utility of the MSSVAE for understanding disease-specific platelet dysfunction mechanisms while identifying universal platelet biomarkers.

\section{Conclusion}

We developed the multi-study sparse deep generative model. The model has spike-and-slab lasso priors that induce sparsity in the mapping from the latent factors to the observed features. Further, the model has sparse mappings corresponding to both a shared latent space among studies, and study-specific latent spaces, allowing for the latent spaces to be interpreted based on the observed features. We prove that the multi-study sparse DGM is identifiable, under an anchor feature assumption. To fit the multi-study sparse DGM, we develop the MSSVAE, both for Gaussian and negative-binomial valued data. On synthetic data, we show the MSSVAE can recover the true latent factors, as well as the feature-factor associations. Further, MSSVAE successfully decomposes platelet transcriptomes into shared universal pathways active across disease states and disease-specific transcriptomic dysregulation. This framework enables identification of biomarkers and therapeutic targets specific to individual diseases while recognizing conserved platelet biology fundamental to hemostasis and thrombosis.

\section*{Acknowledgements}
The authors gratefully acknowledge Rutgers Cyberinfrastructure and AI for Science and Society. A.K. was supported by National Institutes of Health grants 1R21HL181649-01, 1K08HG010061-01A1 and 3UL1TR001085-04S1. 

This work used Bridges-2 at Pittsburgh Supercomputing Center \citep{brown2021bridges} through allocation MTH240044 from the Advanced Cyberinfrastructure Coordination Ecosystem: Services \& Support (ACCESS) program, which is supported by National Science Foundation grants \#2138259, \#2138286, \#2138307, \#2137603, and \#2138296.

\bibliography{bib/references.bib}
\appendix
\section{Proofs}\label{sec:proofs}

\subsection{Proof of \Cref{thm:anchor-id}}

In this section, we prove \Cref{thm:anchor-id}. This requires the following lemmas.

\begin{lemma}\label{lem:parallel}
Under \Cref{ass:anchor,ass:non-parallel,ass:cov}, the anchor features are identifiable from the marginal correlation matrix of the data, $R\defeq \mathrm{Cor}(\bm{x}_i)$. Specifically, the off-diagonal rows $R_{l, \backslash \{l, p\}}$ and $R_{p, \backslash \{l, p\}}$ are parallel if and only if $l$ and $p$ are anchor features for the same factor.
\end{lemma}

\begin{proof}
We first prove that if we have anchor features, then their corresponding rows of the correlation matrix are parallel. For anchor features $j(k), j'(k)$, we have $f_{j'(k)}(z_{ik}) = c_{j'(k)}f_{j(k)}(z_{ik})$ for some $c_{j'(k)}\neq 0$. For any other feature $l$, the noise $\varepsilon_{il}$ is independent of $\bm{z}_i, \varepsilon_{i,j(k)}, \varepsilon_{i, j'(k)}$. Then
for all $l\neq j(k),j'(k)$:
\begin{align*}
\mathrm{Cov}(x_{i,j(k)}, x_{i, l}) &= \mathbb{E}[f_{j(k)}({z_{ik}})f_l(\bm{z}_i)]  - \mathbb{E}[f_{j(k)}(z_{ik})]\mathbb{E}[f_l(\bm{z}_i)] \\
&= \frac{1}{c_{j'(k)}}\left[\mathbb{E}[f_{j'(k)}({z_{ik}})f_l(\bm{z}_i)]  - \mathbb{E}[f_{j'(k)}(z_{ik})]\mathbb{E}[f_l(\bm{z}_i)]\right] \\
&=  \frac{1}{c_{j'(k)}} \mathrm{Cov}(x_{i, j'(k)}, x_{i,l}). 
\end{align*}

As the rows of the covariance matrix (excluding diagonals) are parallel, then the rows $R_{j(k), \backslash j(k), j'(k)}$, $R_{j'(k), \backslash j(k), j'(k)}$ are also parallel. 

Next, we assume $R_{l, \backslash \{l, p\}}$ and $R_{p, \backslash \{l, p\}}$ are parallel, and show that then $l$ and $p$ are anchor features for the same factor.

 Then, there is a $c_{lp}\in\mathbb{R}$ such that for all $r\in \{1,\dots, G\} \backslash \{l, p\}$, 
\begin{align*} 
\mathrm{Cov}(x_{i,l}, x_{i, r}) = c_{lp}\mathrm{Cov}(x_{i,p}, x_{i, r}) \\
\mathrm{Cov}((f(\bm{z}_i)_l - c_{lp}f(\bm{z}_i)_p), x_{i,r}) = 0.
\end{align*}
By \Cref{ass:cov}, then 
\begin{align*}
f(\bm{z}_i)_l = c_{lp}f(\bm{z}_i)_p \quad \text{almost everywhere.}
\end{align*}

By \Cref{ass:non-parallel}, the only functions that are parallel are anchor features. 

Consequently, using the arguments of \citet{bing2023detecting}, we can detect the anchor features from the marginal population correlation matrix. The dimension of the latent space can also be identified from the number of sets of parallel rows.\end{proof}

\begin{lemma}\label{lem:anchor-noise}
Under \Cref{ass:anchor,ass:non-parallel,ass:cov}, the noise variances for anchor features are identifiable.
\end{lemma}

\begin{proof}
For two anchor features $j(k), j'(k)$, for any other feature $l\notin \{j(k), j'(k)\}$ with nonzero covariance with $x_{i,j'(k)}$, we have
\begin{align*}
\mathrm{Cov}(x_{i,j(k)}, x_{i,l})/\mathrm{Cov}(x_{i,j'(k)}, x_{i,l})=1/c_{j'(k)},
\end{align*}
so $c_{j'(k)}$ is identifiable.
Then, for 
\begin{align*}
\mathrm{Cov}(x_{i,j(k)}, x_{i,j'(k)}) = c_{j'(k)} \mathrm{Var}(f_{j(k)}(\bm{z}_i)),
\end{align*}
so $\mathrm{Var}(f_{j(k)}(\bm{z}_i))$ is identifiable. Consequently, 
\begin{align*}
\sigma_{j(k)}^2 = \mathrm{Var}(x_{i,j(k)}) - \mathrm{Var}(f_{j(k)}(\bm{z}_i))
\end{align*}
is identifiable. Thus the noise variances corresponding to anchor features are identifiable. \end{proof}

\begin{lemma}\label{lem:anchor-dist}
Suppose we have two solutions $(f, \bm{W}, p_z, \Sigma)$, $(\widehat{f}, \widehat{\bm{W}}, p_{\widehat{z}}, \widehat{\Sigma})$ such that
\begin{align*}
p(\bm{x}_i|\bm{W}, f, \Sigma) = p(\bm{x}_i|\widehat{\bm{W}},\widehat{f},\widehat{\Sigma})
\end{align*}
Then, under \Cref{ass:anchor,ass:non-parallel,ass:cov}, we have $(z_1,\dots, z_K) \stackrel{d}{=} (h_{1}(\widehat{z}_{\pi(1)}), \dots, h_{K}(\widehat{z}_{\pi(K)}))$, where $\pi$ is a permutation and $h_{k}:\mathbb{R}\to\mathbb{R}$ are invertible functions. 
\end{lemma}

\begin{proof}
From \Cref{lem:parallel}, we can identify the anchor features. This is up to permutation: suppose $x_{i,j(k)}$ is an anchor feature for factor dimension $k$. Then we have $w_{j(k), k}\neq0$, $w_{j(k), k'}=0$, $k'\neq k$. For the alternate solution, we have $\widehat{w}_{j(k),\pi(k)}\neq 0$, $\widehat{w}_{j(k), \pi(k')}=0$, $k'\neq k$, for some permutation $\pi(k)$.

Denote:
\begin{align*}
\bm{V}_i &\defeq (f_{j(1)}(z_{i1}),\dots, f_{j(K)}(z_{iK})),\\
\widehat{\bm{V}}_i &\defeq (\widehat{f}_{j(1)}(\widehat{z}_{i,\pi(1)}), \dots, \widehat{f}_{j(K)}(\widehat{z}_{i, \pi(K)})).
\end{align*}

We have:
\begin{align*}
\bm{V}_i + \bm{\varepsilon}_{i,\mathcal{A}} \stackrel{d}{=} \widehat{\bm{V}}_i + \widehat{\bm{\varepsilon}}_{i,\mathcal{A}}
\end{align*}
where $\bm{\varepsilon}_{i,\mathcal{A}},\widehat{\bm{\varepsilon}}_{i,\mathcal{A}}\stackrel{ind}{\sim} N(0, \Sigma_{\mathcal{A}})$ and $\Sigma_{\mathcal{A}} = \text{diag}(\sigma_{j(1)}^2,\dots, \sigma_{j(K)}^2)$. From \Cref{lem:anchor-noise}, $\Sigma_{\mathcal{A}}$ is identifiable. 

Because the characteristic function of a Gaussian is never zero, we can deconvolve the noise and obtain:
\begin{align*}
 (f_{j(1)}(z_{i1}),\dots, f_{j(K)}(z_{iK})) \stackrel{d}{=} (\widehat{f}_{j(1)}(\widehat{z}_{i,\pi(1)}), \dots, \widehat{f}_{j(K)}(\widehat{z}_{i, \pi(K)})).
\end{align*}

Define the vector map $H:\mathbb{R}^K\to\mathbb{R}^K$, $H(u_1,\dots, u_K)=(f_{j(1)}^{-1}\circ \widehat{f}_{j(1)}(u_1),\dots, f_{j(K)}^{-1}\circ \widehat{f}_{j(K)}(u_K))$. Define $h_k\defeq f_{j(k)}^{-1} \circ \widehat{f}_{j(k)}$.

Then, we have
\begin{align*}
(z_{i1},\dots, z_{iK}) \stackrel{d}{=} (h_1(\widehat{z}_{i,\pi(1)}), \dots, h_K(\widehat{z}_{i, \pi(K)})).
\end{align*}\end{proof}

\begin{lemma}\label{lem:cond-mean-ae}
Suppose we have two solutions $(f, \bm{W}, p_z, \Sigma)$, $(\widehat{f}, \widehat{\bm{W}}, p_{\widehat{z}}, \widehat{\Sigma})$ such that
\begin{align*}
p(\bm{x}|\bm{W}, f, \Sigma) = p(\bm{x}|\widehat{\bm{W}},\widehat{f},\widehat{\Sigma}).
\end{align*}
Define $\bm{V}_i$ as in \Cref{lem:anchor-dist}. For $l=1,\dots, G$, define:
\begin{align*}
{g}_l(v_1,\dots, v_K) \defeq {f}_l({w}_{l1}{f}_{j(1)}^{-1}(v_1),\dots, {w}_{lK}{f}_{j(K)}^{-1}(v_K))
\end{align*}
and
\begin{align*}
\widehat{g}_l({v}_1,\dots, {v}_K) \defeq \widehat{f}_l(\widehat{w}_{l1}\widehat{f}_{j(1)}^{-1}({v}_1),\dots, \widehat{w}_{lK}\widehat{f}_{j(K)}^{-1}({v}_K)).
\end{align*}
Then, with \Cref{ass:anchor,ass:non-parallel,ass:cov}, for all features $x_{il}$, $l=1,\dots, G$, we have
\begin{align*}
g_l(\bm{v}) = \widehat{g}_l(\bm{v})\text{ a.e.} P_{\bm{V}}.
\end{align*}
  
\end{lemma}
\begin{proof}
Let $\bm{X}_{i,\mathcal{A}} = (\bm{x}_{i,j(1)},\dots, \bm{x}_{i,j(K)})$ be the vector of anchor features. Let $\bm{X}_{i,\backslash\mathcal{A}}$ be the vector of the  non-anchor features.

To prove this lemma, we need to identify the joint distribution $P_{\bm{V}, \bm{X}_{\backslash \mathcal{A}}}$.

Considering the characteristic functions:
\begin{align*}
\varphi_{\bm{X}_{\mathcal{A}},\bm{X}_{\backslash\mathcal{A}}}(t, s) &= \varphi_{\bm{V},\bm{X}_{\backslash\mathcal{A}}}(t, s)\varphi_{\varepsilon_{\mathcal{A}}}(t). 
\end{align*}
So,
\begin{align*}
\varphi_{\bm{V},\bm{X}_{\backslash\mathcal{A}}}(t, s)   = \frac{\varphi_{\bm{X}_{\mathcal{A}},\bm{X}_{\backslash\mathcal{A}}}(t, s)}{\varphi_{\varepsilon_{\mathcal{A}}}(t)}  
\end{align*}
is identified (Gaussian characteristic function is non-zero).

For anchor features, we have $g_{j(k)}(\bm{v}) = v_k = \widehat{g}_{j(k)}(\bm{v})$ a.e. $P_{\bm{V}}$ by definition.

For non-anchor features, we then have:
\begin{align*}
\mathbb{E}[x_{il}|\bm{V}=\bm{v}] &= g_l(\bm{v}) \\
&=\widehat{g}_l(\bm{v}) \text{ a. e. on } P_{\bm{V}}
\end{align*}
because the joint distribution of $(\bm{X}_{\backslash \mathcal{A}}, \bm{V})$ is identifiable.

\end{proof}

\begin{lemma}\label{lem:w}
Suppose we have two solutions $(f, \bm{W}, p_z, \Sigma)$, $(\widehat{f}, \widehat{\bm{W}}, p_{\widehat{z}}, \widehat{\Sigma})$ such that
\begin{align*}
p(\bm{x}|\bm{W}, f, \Sigma) = p(\bm{x}|\widehat{\bm{W}},\widehat{f},\widehat{\Sigma})
\end{align*}

Then, with \Cref{ass:anchor,ass:non-parallel,ass:cov,ass:cond-non-degen,ass:fun-non-degen}, the support of $\bm{W}$ and $\widehat{\bm{W}}$ is the same (up to permutation of columns).
\end{lemma}
\begin{proof}
By relabeling the columns of \(\widehat{\bm W}\), we assume without loss
of generality that the permutation \(\pi\) in Lemma~\ref{lem:anchor-dist}
is the identity.

From \Cref{lem:cond-mean-ae}, we have
\begin{align*}
g_l(v_{i1},\dots, v_{iK}) = \widehat{g}_{l}(v_{i1},\dots, v_{iK}) \text{ a.e.} P_{\bm{V}}.
\end{align*}

We argue by contradiction. Suppose $w_{l1}=0$ but $\widehat{w}_{l1}\neq 0$.

If \(w_{l1}=0\), then
\[
\operatorname{Var}\{g_l(\bm V)\mid \bm V_{\setminus 1}\}=0
\quad\text{a.s.}
\]
Since \(g_l(\bm V)=\widehat g_l(\bm V)\) a.e., the same must hold for
\(\widehat g_l\). This contradicts functional non-degeneracy whenever
\(\widehat w_{l1}\neq 0\) (\Cref{ass:cond-non-degen,ass:fun-non-degen}).

We repeat this argument for all $l=1,\dots, G$, $k=1,\dots, K$.
Consequently, we must have $\mathrm{supp}(\widehat{\bm{W}})\subseteq \mathrm{supp}({\bm{W}})$. We can make the same argument in reverse to obtain $\mathrm{supp}({\bm{W}})\subseteq\mathrm{supp}(\widehat{\bm{W}})$. 
\end{proof}

\begin{lemma}\label{lem:noise-all}
The noise variances $\Sigma$ are identifiable.
\end{lemma}

\begin{proof}
We need to prove that the noise variance of the non-anchor features are identifiable.

Now,
\begin{align*}
\text{Var}(x_{l}) &= \text{Var}(\mathbb{E}[x_{l}|\bm{V}]) + \mathbb{E}[\text{Var}(x_{l}|\bm{V})] \\
&= \sigma_l^2 + \text{Var}(g_l(v_1,\dots, v_K)) \\
&= \sigma_l^2 + \text{Var}(\widehat{g}_l(v_1,\dots, v_K)) \quad (\Cref{lem:cond-mean-ae}) \\
&= \widehat{\sigma}_l^2+ \text{Var}(\widehat{g}_l(v_1,\dots, v_K)) \quad \text{(equality of distributions)}.
\end{align*}

Hence, $\sigma_l^2 = \widehat{\sigma}_l^2$.\end{proof}

Together, these lemmas prove \Cref{thm:anchor-id}.

\begin{proof}
Combining \Cref{lem:parallel,lem:anchor-noise,lem:anchor-dist,lem:w,lem:noise-all}, the theorem is proved. \end{proof}

\subsection{Proof of \Cref{thm:nb-id}}

In this section, we prove  the negative-binomial case (\Cref{thm:nb-id}).

\begin{lemma}\label{lem:nb-parallel}
Under \Cref{ass:nb-anchor,ass:non-parallel,ass:cov}, the anchor features are identifiable from the correlation matrix of the data normalized by the library size, $R\defeq \mathrm{Cor}(\bm{x}_i/\ell_i)$. Specifically, the off-diagonal rows $R_{l, \backslash \{l, p\}}$ and $R_{p, \backslash \{l, p\}}$ are parallel if and only if $l$ and $p$ are anchor features for the same factor.
\end{lemma}

\begin{proof}
We first prove that if we have anchor features, then their corresponding rows of the correlation matrix are parallel. Let $y_{ij} = x_{ij}/\ell_i$. For anchor features $j(k), j'(k)$, we have $f_{j'(k)}(z_{ik}) = c_{j'(k)}f_{j(k)}(z_{ik})$ for some $c_{j'(k)}\neq 0$. Given \(\bm z_i\), the counts
\(x_{i1},\ldots,x_{iG}\) are conditionally independent. Hence, 
for all $l\neq j(k),j'(k)$:
\begin{align*}
\mathrm{Cov}(y_{i,j(k)}, y_{i, l}) &= \mathbb{E}[f_{j(k)}({z_{ik}})f_l(\bm{z}_i)]  - \mathbb{E}[f_{j(k)}(z_{ik})]\mathbb{E}[f_l(\bm{z}_i)] \\
&= \frac{1}{c_{j'(k)}}\left[\mathbb{E}[f_{j'(k)}(\bm{z_{ik}})f_l(\bm{z}_i)]  - \mathbb{E}[f_{j'(k)}(z_{ik})]\mathbb{E}[f_l(\bm{z}_i)]\right] \\
&=  \frac{1}{c_{j'(k)}} \mathrm{Cov}(y_{i, j'(k)}, y_{i,l}). 
\end{align*}

As the rows of the covariance matrix (excluding diagonals) are parallel, then the rows $R_{j(k), \backslash j(k), j'(k)}$, $R_{j'(k), \backslash j(k), j'(k)}$ are also parallel. 

Next, we assume $R_{l, \backslash \{l, p\}}$ and $R_{p, \backslash \{l, p\}}$ are parallel, and show that then $l$ and $p$ are anchor features for the same factor.

 Then, there is a $c_{lp}\in\mathbb{R}$ such that for all $r\in \{1,\dots, G\} \backslash \{l, p\}$, 
\begin{align*} 
\mathrm{Cov}(y_{i,l}, y_{i, r}) = c_{lp}\mathrm{Cov}(y_{i,p}, y_{i, r}) \\
\mathrm{Cov}(f_l(\bm{z}_i) - c_{lp}f_p(\bm{z}_i),  f_r(\bm{z}_i)) = 0.
\end{align*}
By \Cref{ass:cov}, then 
\begin{align*}
f(\bm{z}_i)_l = c_{lp}f(\bm{z}_i)_p \quad \text{almost everywhere.}
\end{align*}

By \Cref{ass:non-parallel}, the only functions that are parallel are anchor features. 

Consequently, using the arguments of \citet{bing2023detecting}, we can detect the anchor features from the normalized population correlation matrix. The dimension of the latent space can also be identified from the number of sets of parallel rows.\end{proof}

\begin{lemma}\label{lem:nb-disp-id}
For the anchor features $\mathcal{A}$, the inverse-dispersion parameters $\phi_{\mathcal{A}} = \{\phi_j: j \in \mathcal{A}\}$ are identifiable.
\end{lemma}

\begin{proof}
We have for any feature $l\notin\{j(k), j'(k)\}$ with nonzero covariance with $x_{i,j'(k)}$,
\begin{align*}
\mathrm{Cov}(x_{i,j(k)}, x_{i,l}) /\mathrm{Cov}(x_{i,j'(k)}, x_{i,l}) = 1/c_{j'(k)},
\end{align*}
so $c_{j'(k)}$ is identifiable.

Then, $\mathrm{Cov}(x_{i,j(k)}, x_{i,j'(k)}) = c_{j'(k)} \ell_i^2 \mathrm{Var}(f_{j(k)}(z_{ik}))$, so $\mathrm{Var}(f_{j(k)}(z_{ik}))$ is identifiable. 

We have:
\begin{align*}
\operatorname{Var}(x_{i,j(k)}) &= \operatorname{Var}(\mathbb{E}[x_{i,j(k)}|\bm{z}_i]) + \mathbb{E}[\operatorname{Var}(x_{i,j(k)}|\bm{z}_i)] \\
&= \ell_i^2\operatorname{Var}(f_{j(k)}(z_{ik})) + \ell_i\mathbb{E}[ f_{j(k)}(z_{ik})] + \frac{\ell_i^2 \mathbb{E}[(f_{j(k)}(z_{ik}))^2]}{\phi_{j(k)}}
\end{align*}

So
\begin{align*}
\phi_{j(k)} = \frac{\ell_i^2\mathbb{E}[( f_{j(k)}(z_{ik}))^2]}{\operatorname{Var}(x_{i,j(k)}|\ell_i) - \ell_i^2\operatorname{Var}( f_{j(k)}(z_{ik})) -\ell_i \mathbb{E}[ f_{j(k)}(z_{ik})]}.
\end{align*}

Now,  $\ell_i\mathbb{E}[ f_{j(k)}(z_{ik})] = \mathbb{E}[ x_{i,j(k)}]$.  Thus, the inverse-dispersion parameters for the anchor features are identifiable. 

\end{proof}

\begin{lemma}\label{lem:nb-noise-rm}
Define
\begin{align*}
\bm V_i &= (f_{j(1)}(z_{i1}),\dots,f_{j(K)}(z_{iK})) \\
\mathcal{A} &= \{j(k): k= 1,\dots, K\}
\end{align*}
For the negative-binomial model \Cref{eq:nb-svae} with $\ell_i$  known and fixed, the distribution of $\bm{V}_i$ is identifiable from the marginal distribution of $\bm{X}_{\mathcal{A}}$, denoted $P_{\bm{X}_{\mathcal{A}}}$.
\end{lemma}

\begin{proof}
We have:
\begin{align*}
x_{ij}|m_{ij} \sim \mathrm{NB}(\mathrm{mean}=m_{ij}, \mathrm{inv}\text{-}\mathrm{disp}=\phi_j),
\end{align*}
where $m_{ij} = \ell_if_j(\bm{w}_j\odot\bm{z}_i)$ and $\ell_i > 0$ is fixed and known.

We show the joint distribution of $\bm{M}_{i,\mathcal{A}}$ is identifiable from the marginal distribution $P_{\bm{X}_{\mathcal{A}}}$.

First, we reparameterize the NB model with:
\begin{align}
r_{ij} = \frac{m_{ij}}{m_{ij} + \phi_j} \in [0,1)
\end{align}
Let $G$ be the distribution of $(r_{i1},\dots, r_{i,j(K)})$. Then the marginal distribution is:
\begin{align*}
P(\bm{X}_{i, \mathcal{A}}=\bm{x}) &= \prod_{j=1}^K\frac{\Gamma(x_j+\phi_j)}{\Gamma(\phi_j)x_j!}\int_{[0,1)^G} \prod_{j=1}^K r_j^{x_j}(1-r_j)^{\phi_j} dG(\bm{r})  \\
&=  \prod_{j=1}^K\frac{\Gamma(x_j+\phi_j)}{\Gamma(\phi_j)x_j!}\int_{[0,1)^K} \prod_{j=1}^K r_j^{x_j} dH(\bm{r})
\end{align*}
where $dH(\bm{r}) = \prod_{j=1}^K (1-r_j)^{\phi_j}dG(\bm{r})$. Note that we used conditional independence of $X_{i1},\dots, X_{i,j(K)}$ given $\bm{M}_{i,\mathcal{A}}$.

This is a power-series distribution. Consequently, we can obtain all moments of $H$ from the observed distribution of $\bm{X}_i$.  A measure on $[0,1]^K$ is uniquely determined by its moments, so $H$ is identifiable (multivariate Hausdorff moment determinacy, \citet{sapatinas1995identifiability}). Then, $G$ is identifiable from $H$.   Note this argument requires $\phi_j$ to be known, as we obtain from \Cref{lem:nb-disp-id}.

As $r_{ij}$ is a one-to-one transformation of $m_{ij}$, we have the distribution of $\bm{M}_{i,\mathcal{A}}$ is identifiable from $P_{\bm{X}_{\mathcal{A}}}$. Then, the distribution of $\bm{V}_i = \bm{M}_{i,\mathcal{A}}/\ell_i$ is identifiable as $\ell_i>0$ is known.

\end{proof}

\begin{lemma}\label{lem:nb-anchor-dist}
Suppose we have two solutions $(f, \bm{W}, p_z,\phi)$, $(\widehat{f}, \widehat{\bm{W}}, p_{\widehat{z}},\widehat{\phi})$ such that
\begin{align*}
p(\bm{x}_i|\bm{W}, f, \phi) = p(\bm{x}_i|\widehat{\bm{W}},\widehat{f},\widehat{\phi})
\end{align*}
Then, under \Cref{ass:nb-anchor,ass:non-parallel,ass:cov}, we have $(z_1,\dots, z_K) \stackrel{d}{=} (h_{1}(\widehat{z}_{\pi(1)}), \dots, h_{K}(\widehat{z}_{\pi(K)}))$, where $\pi$ is a permutation and $h_{k}:\mathbb{R}\to\mathbb{R}$ are invertible functions. 
\end{lemma}

\begin{proof}
From \Cref{lem:nb-parallel}, we can identify the anchor features. This is up to permutation: suppose $x_{i,j(k)}$ is an anchor feature for factor dimension $k$. Then we have $w_{j(k), k}\neq0$, $w_{j(k), k'}=0$, $k'\neq k$. For the alternate solution, we have $\widehat{w}_{j(k),\pi(k)}\neq 0$, $\widehat{w}_{j(k), \pi(k')}=0$, $k'\neq k$, for some permutation $\pi(k)$.

Denote:
\begin{align*}
\bm{V}_i &\defeq (f_{j(1)}(z_{i1}),\dots, f_{j(K)}(z_{iK})),\\
\widehat{\bm{V}}_i &\defeq (\widehat{f}_{j(1)}(\widehat{z}_{i,\pi(1)}), \dots, \widehat{f}_{j(K)}(\widehat{z}_{i, \pi(K)})).
\end{align*}

From \Cref{lem:nb-noise-rm}, we have
\begin{align*}
\bm{V}_i  \stackrel{d}{=} \widehat{\bm{V}}_i. 
\end{align*}

Define the vector map $H:\mathbb{R}^K\to\mathbb{R}^K$, $H(u_1,\dots, u_K)=(f_{j(1)}^{-1}\circ \widehat{f}_{j(1)}(u_1),\dots, f_{j(K)}^{-1}\circ \widehat{f}_{j(K)}(u_K))$. Define $h_k\defeq f_{j(k)}^{-1} \circ \widehat{f}_{j(k)}$. (Note these exist from anchor monotonicity, \Cref{ass:nb-anchor}). 

Then, we have
\begin{align*}
(z_{i1},\dots, z_{iK}) \stackrel{d}{=} (h_1(\widehat{z}_{i,\pi(1)}), \dots, h_K(\widehat{z}_{i, \pi(K)})).
\end{align*}
\end{proof}

\begin{lemma}\label{lem:nb-cond-mean-ae}
Suppose we have two solutions $(f, \bm{W}, p_z, \bm{\phi})$, $(\widehat{f}, \widehat{\bm{W}}, p_{\widehat{z}}, \widehat{\phi})$ such that
\begin{align*}
p(\bm{x}|\bm{W}, f, \phi,\ell) = p(\bm{x}|\widehat{\bm{W}},\widehat{f},\widehat{\bm{\phi}},\ell).
\end{align*}
Define $\bm{V}_i$ as in \Cref{lem:anchor-dist}. For $l=1,\dots, G$, define:
\begin{align*}
{g}_l(v_1,\dots, v_K) \defeq {f}_l({w}_{l1}{f}_{j(1)}^{-1}(v_1),\dots, {w}_{lK}{f}_{j(K)}^{-1}(v_K))
\end{align*}
and
\begin{align*}
\widehat{g}_l({v}_1,\dots, {v}_K) \defeq \widehat{f}_l(\widehat{w}_{l1}\widehat{f}_{j(1)}^{-1}({v}_1),\dots, \widehat{w}_{lK}\widehat{f}_{j(K)}^{-1}({v}_K)).
\end{align*}
Then, with \Cref{ass:nb-anchor,ass:non-parallel,ass:cov}, for all features $x_{il}$, $l=1,\dots, G$, we have
\begin{align*}
g_l(\bm{v}) = \widehat{g}_l(\bm{v})\text{ a.e.} P_{\bm{V}}.
\end{align*}
\end{lemma}

\begin{proof}
We first show that the joint distribution of $(\bm{V}, \bm{X}_{\backslash \mathcal{A}})$ is identifiable.  

For each value
\[
\bm y\in \mathbb N^{G-|\mathcal A|}
\]
of \(\bm X_{\backslash\mathcal A}\), define the finite measure
\[
G_{\bm y}(B)
=
P(\bm V\in B,\bm X_{\backslash\mathcal A}=\bm y),
\qquad B\subseteq \mathbb R_+^K.
\]

We have:
\begin{align*}
P(\bm{X}_{\mathcal{A}}=\bm{x}_{\mathcal{A}}, \bm{X}_{\backslash \mathcal{A}}=\bm{y})
&=\int P(\bm{X}_{\mathcal{A}}=\bm{x}_{\mathcal{A}}|\bm{V}=\bm{v})  dG_{\bm{y}}(\bm{v}),
\end{align*}
where we used that $\bm{X}_{\mathcal{A}}\indep \bm{X}_{\backslash \mathcal{A}}|\bm{V}$. This conditional independence holds because $\bm{Z}$ is determined by $\bm{V}$ coordinate-wise (due to strict monotonicity of anchor maps), and the features are conditionally independent given $\bm{Z}$.

For each fixed value $\bm{y}$, we have $P(\bm{X}_{\mathcal{A}} =\bm{x}_{\mathcal{A}}| \bm{V}=\bm{v})$ is the multivariate negative-binomial kernel, so the measure $G_{\bm{y}}(\bm{v})$ is identified by the moment argument of \Cref{lem:nb-noise-rm}. As this holds for every value $\bm{y}$ of $\bm{X}_{\backslash \mathcal{A}}$,  the joint distribution $P_{\bm{V}, \bm{X}_{\backslash \mathcal{A}}}$ is identifiable.

For anchor features, we have $g_{j(k)}(\bm{v}) = v_k = \widehat{g}_{j(k)}(\bm{v})$ a.e. $P_{\bm{V}}$ by definition.

For non-anchor features, we then have:
\begin{align*}
\mathbb{E}[x_{il}|\bm{V}=\bm{v}] &= \ell_i g_l(\bm{v}) \\
&=\ell_i \widehat{g}_l(\bm{v}) \text{ a. e. on } P_{\bm{V}}
\end{align*}
because the joint distribution of $(\bm{X}_{\backslash \mathcal{A}}, \bm{V})$ is identifiable.

\end{proof}

\begin{lemma}\label{lem:nb-w}
Suppose we have two solutions $(f, \bm{W}, p_z, \bm{\phi})$, $(\widehat{f}, \widehat{\bm{W}}, p_{\widehat{z}},\widehat{\bm{\phi}})$ such that
\begin{align*}
p(\bm{x}|\bm{W}, f,\bm\phi,\ell) = p(\bm{x}|\widehat{\bm{W}},\widehat{f},\widehat{\bm\phi},\ell)
\end{align*}

Then, with \Cref{ass:nb-anchor,ass:non-parallel,ass:cov,ass:cond-non-degen,ass:fun-non-degen}, the support of $\bm{W}$ and $\widehat{\bm{W}}$ is the same (up to permutation of columns).

\end{lemma}

\begin{proof}
By relabeling the columns of \(\widehat{\bm W}\), we assume without loss
of generality that the permutation \(\pi\) in \Cref{lem:nb-anchor-dist}
is the identity.

From \Cref{lem:nb-cond-mean-ae}, we have
\begin{align*}
\mathbb{E}[x_{ij}|\bm{V}=\bm{v}] &= \ell_i g_j(v_1,\dots, v_K) \\
&= \ell_i \widehat{g}_j(v_1,\dots, v_K) \text{ a.e. on } P_{\bm{V}|L}.
\end{align*}

We argue by contradiction. Suppose $w_{l1}=0$ but $\widehat{w}_{l1}\neq 0$.

If \(w_{l1}=0\), then
\[
\operatorname{Var}\{g_l(\bm V)\mid \bm V_{\setminus 1}\}=0
\quad\text{a.s.}
\]
Since \(g_l(\bm V)=\widehat g_l(\bm V)\) a.e., the same must hold for
\(\widehat g_l\). This contradicts functional non-degeneracy whenever
\(\widehat w_{l1}\neq 0\) (\Cref{ass:cond-non-degen,ass:fun-non-degen}).

We repeat this argument for all $l=1,\dots, G$, $k=1,\dots, K$.
Consequently, we must have $\mathrm{supp}(\widehat{\bm{W}})\subseteq \mathrm{supp}({\bm{W}})$. We can make the same argument in reverse to obtain $\mathrm{supp}({\bm{W}})\subseteq\mathrm{supp}(\widehat{\bm{W}})$. 

\end{proof}

\begin{lemma}
The inverse-dispersion parameters are identifiable.
\end{lemma}

\begin{proof}
For the non-anchor features, we have:
\begin{align*}
\mathrm{Var}(x_{ij}) &= \mathbb{E}[\mathrm{Var}(x_{ij}|\bm{V}_i)] + \mathrm{Var}(\ell_i g_j(\bm{V}_i)) \\
&= \mathbb{E}[\ell_i g_{j}(\bm{V}_i)] + \frac{\mathbb{E}[(\ell_i g_j(\bm{V}_i))^2]}{\phi_j} +\mathrm{Var}(\ell_i g_j(\bm{V}_i)).
\end{align*}
As $g_j(\bm{v}) = \hat{g}_j(\bm{v})$ a.e. on $P_{\bm{V}}$, $\phi_j$ is identifiable.

\end{proof}

Combining these lemmas, \Cref{thm:nb-id} is proved.

\subsection{Proof of \Cref{thm:shared-id}}
Now, we prove \Cref{thm:shared-id}.

\begin{proof}
First, we consider each study separately. From \Cref{thm:anchor-id}, for each study $m$, we can identify the support of $[\bm{W}^{(S)}, \bm{W}^{(m)}]$ up to column permutation. (Note: column permutation means that the shared and study-specific columns cannot be partitioned). 

Second, we consider all studies and extract the columns of $\{[\bm{W}^{(S)}, \bm{W}^{(m)}]\}_{m=1}^M$ which have the same support across all studies. These columns form $\bm{W}^{(S)}$.

Finally, the remaining columns for each study are the study-specific columns $\bm{W}^{(m)}$. By construction, these columns did not match across all studies (otherwise they would be in $\bm{W}^{(S)}$). Because of the anchor feature structure, the columns $\{\bm{W}^{(m)}\}_{m=1}^M$ have disjoint supports on the anchor indices, and so the intersection of their column spaces is null (i.e. they do not have shared signal remaining). 
\end{proof}

\section{Implementation details}\label{sec:comp-det}

\subsection{Architecture}

For the encoders, we use two-layer ReLU feed forward neural networks with batch norm:
\begin{align*}
\bm{h}_\psi^{(1)}(\bm{x}_i) &= \mathrm{ReLU}(\mathrm{BN}_\psi^{(1)}(\bm{V}_\psi^{(1)}\bm{x}_i + \bm{b}_\psi^{(1)})) \\
\bm{h}_\psi^{(2)}(\bm{x}_i) &= \mathrm{ReLU}(\mathrm{BN}_\psi^{(2)}(\bm{V}_\psi^{(2)}\bm{h}_\psi^{(1)}(\bm{x}_i) + \bm{b}_e^{(2)})) \\
\mu_{\psi}(\bm{x}_i) &= \bm{V}_\psi^{(\mu)}\bm{h}_\psi^{(2)}(\bm{x}_i) + \bm{b}_\psi^{(\mu)} \\
\sigma_{\psi}(\bm{x}_i) &= \bm{V}_\psi^{(\sigma)}\bm{h}_\psi^{(2)}(\bm{x}_i) + \bm{b}_\psi^{(\sigma)},
\end{align*}
where $\bm{V}_{\psi}^{(l)}\in\mathbb{R}^{d_\mathrm{out}\times d_{\mathrm{in}}}$ is the weight matrix for layer $l$ with input dimension $d_{\mathrm{in}}$ and output dimension $d_{\mathrm{out}}$, and $\bm{b}_{\psi}^{(l)}\in\mathbb{R}^{d_\mathrm{out}}$ is the bias/intercept vector for layer $l$.   Further, $\mathrm{BN}^{(l)}$ denotes standard batch norm \citep{ioffe2015batch} and $\mathrm{ReLU}(z) = \max\{0, z\}.$

For the decoder, we use a two-layer ReLU feed forward neural network with a skip connection from the masked latents to the output dimension. Specifically, for the output dimension $j$ of sample $i$ from study $m$, denote the masked latents as $\bm{c}_{ij}^{(m)} = \widetilde{\bm{w}}_j^{(m)}\odot \widetilde{\bm{z}}_i^{(m)}$. Then,
\begin{align*}
\bm{h}_{\theta}^{(1)}(\bm{c}_{ij}^{(m)}) &= \mathrm{ReLU}(\bm{V}_{\theta}^{(1)}(\bm{c}_{ij}^{(m)})) \\
\bm{h}_{\theta}^{(2)}(\bm{c}_{ij}^{(m)}) &= \mathrm{ReLU}(\bm{V}_\theta^{(2)}\bm{h}_{\theta}^{(1)}(\bm{c}_{ij}^{(m)}) + \bm{b}_{\theta}^{(2)}) \\
f_j(\bm{c}_{ij}^{(m)}) &=
\begin{cases}
(\bm{v}_{\theta, j}^{(3)})^\top [\bm{h}_{\theta}^{(2)}(\bm{c}_{ij}^{(m)}), \bm{c}_{ij}^{(m)}] + b_{\theta, j}^{(3)} &\text{(Gaussian MSSVAE)} \\
\mathrm{softplus}[(\bm{v}_{\theta, j}^{(3)})^\top [\bm{h}_{\theta}^{(2)}(\bm{c}_{ij}^{(m)}) , \bm{c}_{ij}^{(m)}] + b_{\theta, j}^{(3)}] &\text{(NB-MSSVAE)}
\end{cases}
\end{align*}
where $\bm{v}_{\theta, j}^\top$ is the $j$th row of $\bm{V}_{\theta}^{(3)}$. That is, each of the $G$ outputs of $f$ share all neural network parameters until the final layer. 

The dimensions of the hidden layers used for experiments are given in \Cref{tab:hyperparam}.

\subsection{Optimization}

The objective function is:
\begin{align}
\mathcal{L}(\Theta) &= \sum_{m=1}^M\left\{\sum_{i=1}^{n_m} \left\{\mathbb{E}_{q_{\psi_S}(\bm{z}_{i}^{(m)}|\bm{x}_i^{(m)})q_{\psi_m}(\bm{\zeta}_{i}^{(m)}|\bm{x}_i^{(m)})}\left[ \log p_{\theta}(\bm{x}_i^{(m)} | \bm{W}^{(S)}, \bm{W}^{(m)}, \bm{z}_i^{(m)}, \bm{\zeta}_i^{(m)}, \bm{\Sigma}) \right] \right.\right. \notag\\
&\quad \left. -  D_{KL}(q_{\psi_S}(\bm{z}_{i}^{(m)}|\bm{x}_i^{(m)}) || p(\bm{z}_{i}^{(m)})) - D_{KL}(q_{\psi_m}(\bm{\zeta}_{i}^{(m)}|\bm{x}_i^{(m)}) || p(\bm{\zeta}_{i}^{(m)})) \right\}\notag\\
&\quad  \left. + \mathbb{E}_{\bm{\Gamma}^{(m)}|\bm{W}^{(m)}, \bm{\eta}^{(m)}}{[\log p(\bm{W}^{(m)}|\bm{\Gamma}^{(m)}) p(\bm{\Gamma}^{(m)}|\bm{\eta}^{(m)})p(\bm{\eta}^{(m)})]}\right\}
  \notag \\
&\quad   +  \mathbb{E}_{\bm{\Gamma}^{(S)}|\bm{W}^{(S)}, \bm{\eta}^{(S)}}{[\log p(\bm{W}^{(S)}|\bm{\Gamma}^{(S)}) p(\bm{\Gamma}^{(S)}|\bm{\eta}^{(S)})p(\bm{\eta}^{(S)})]} + \log p(\bm{\Sigma}). \label{eq:ELBO-gauss-app}
\end{align}

We optimize \Cref{eq:ELBO-gauss-app} by alternating between an expectation step and a maximization step. 

\subsubsection{Expectation step}
To approximate the first term of \Cref{eq:ELBO-gauss-app}, we use Monte Carlo. When the likelihood is Gaussian, this corresponds to:
\begin{align}
\mathbb{E}_{q_{\psi_S}(\bm{z}_{i}^{(m)}|\bm{x}_i^{(m)})q_{\psi_m}(\bm{\zeta}_{i}^{(m)}|\bm{x}_i^{(m)})}&\left[ \log p_{\theta}(\bm{x}_i^{(m)} | \bm{W}^{(S)}, \bm{W}^{(m)}, \bm{z}_i^{(m)}, \bm{\zeta}_i^{(m)}, \bm{\Sigma}) \right] \\
&\quad\approx -\frac{1}{2L}\sum_{l=1}^L\sum_{j=1}^G \frac{1}{\sigma_j^2} \left[x_{ij} - f_j(\widetilde{\bm{w}}_j^{(m)} \odot \widetilde{\bm{z}}_i^{(m), (l)})  \right]^2 + \log(2\pi\sigma_j^2), \label{eq:e-step-gauss}
\end{align}
where we draw the samples  $\widetilde{\bm{z}}_i^{(m), (l)}$ as
\begin{align*}
    \widetilde{\bm{z}}_i^{(m), (l)} = 
    \begin{pmatrix}
        \bm{z}_i^{(m),(l)} & \bm{0}_{1\times (K_S+K_1+\cdots + K_{m-1})} & \bm{\zeta}_i^{(m), (l)} & \bm{0}_{1\times (K_{m+1}+\cdots + K_{M})}
    \end{pmatrix}
\end{align*} 
with
\begin{align*}
 \bm{z}_i^{(m),(l)} &= \mu_{\psi_S}(\bm{x}_i) + \sigma_{\psi_S}(\bm{x}_i) \odot \bm{\zeta}_{i}^{(m), (l)}, \quad \bm{\xi}_{i}^{(m), (l)}\sim N(\bm{0},\bm{I}),\\
 \bm{\zeta}_i^{(m),(l)} &= \mu_{\psi_m}(\bm{x}_i) + \sigma_{\psi_m}(\bm{x}_i) \odot \bm{\nu}_{i}^{(m), (l)}, \quad \bm{\nu}_{i}^{(m), (l)}\sim N(\bm{0},\bm{I}).
\end{align*}
That is, we draw standard Gaussian random variables and then obtain samples from the variational posterior by multiplying by standard deviation and adding the mean. For the NB-MSSVAE, the process is the same, replacing \Cref{eq:e-step-gauss} with the negative-binomial log likelihood.

For the remaining terms, the calculations are the same for the shared and study-specific parameters, so we write the general formula, omitting the superscripts.

The KL divergence between the variational posterior $q_{\psi}(\bm{z}_i|\bm{x}_i)$ and the prior $\bm{z}_i\sim N(\bm{0}, \bm{I})$ is:
\begin{align*}
D_{KL}(q_{\psi}(\bm{z}_i|\bm{x}_i)|| p(\bm{z}_i)) = -\frac{1}{2}\sum_{k=1}^K \left[1 + \log(\sigma_{\psi}^2(\bm{x}_i)) - (\mu_{\psi}(\bm{x}_i))^2 - \sigma_{\psi}^2(\bm{x}_i)\right].
\end{align*}

For the spike-and-slab lasso prior terms, we have:
\begin{align*}
\mathbb{E}_{\bm{\Gamma}|\bm{W}^{\mathrm{old}}, \bm{\eta}^{\mathrm{old}}}{[\log p(\bm{W}|\bm{\Gamma}) p(\bm{\Gamma}|\bm{\eta})p(\bm{\eta})]} &= -\sum_{k=1}^K\sum_{j=1}^G \lambda^*(w_{jk}^{\mathrm{old}},\eta_k^{\mathrm{old}})|w_{jk}| \\
&\quad + \sum_{k=1}^K\left[ \sum_{j=1}^G \mathbb{E}[\gamma_{jk}|w_{jk}^{\mathrm{old}}, \eta_k^{\mathrm{old}}] + a - 1\right]\log\eta_k \\
&\qquad+ \left[G-\sum_{j=1}^G \mathbb{E}[\gamma_{jk}|w_{jk}^{\mathrm{old}}, \eta_k^{\mathrm{old}}]  + b - 1\right]\log(1-\eta_k),
\end{align*}
where 
\begin{align*}
\mathbb{E}[\gamma_{jk}|w_{jk}, \eta_k] &= \frac{\eta_k \lambda_1 \exp(-\lambda_1|w_{jk}|)}{\eta_k\lambda_1\exp(-\lambda_1|w_{jk}|) + (1-\eta_k)\lambda_0\exp(-\lambda_0|w_{jk}|)} \\
\lambda^*(w_{jk},\eta_k) &= \lambda_1\mathbb{E}[\gamma_{jk}|w_{jk}, \eta_k]  + \lambda_0(1-\mathbb{E}[\gamma_{jk}|w_{jk}, \eta_k] ).
\end{align*}

\subsubsection{Maximization step}

In the maximization step, we take a gradient step over parameters $\Theta$. The derivatives of $L(\bm{\Theta})$ are calculated using automatic differentiation.

\subsubsection{Stochastic optimization}

In experiments, we use stochastic optimization. At each epoch, we split our data into mini-batches. For each mini-batch, we evaluate the expectations and take a gradient step. For the
gradient steps, we use Adam \citep{kingma2015adam} with the default PyTorch parameters $(\beta_1=0.9,\beta_2=0.999)$.

The mini-batches are selected using PyTorch's \texttt{WeightedRandomSampler} with each sample assigned a weight inversely proportional to the size of its associated group. This ensures each mini-batch has an approximately equal number of samples from each group.

\section{Empirical study details}\label{sec:emp-det}

\subsection{Metrics}
In this section, we provide more details about the metrics used for evaluation in the simulation studies.

The consensus \citep{H10}, and relevance and recovery scores \citep{P06} are all based on the Jaccard index, a measure of similarity between two sets $A$ and $B$:
\begin{align}
J(A, B) = \frac{|A\cap B|}{|A \cup B|}. \label{eq:jaccard_score}
\end{align}
The Jaccard index penalizes estimates $\widehat{\bm{W}}$ which overestimate the support of the true $\bm{W}$.

Denote $C_k$ as the set non-zero entries of the $k$th column $\bm{W}_{\cdot, k}$. Let $\mathcal{C}_{t} = \{C_1,\dots, C_K\}$ and let $\mathcal{C}_f = \{\widehat{C}_1,\dots, \widehat{C}_{\widehat{K}}\}$, where $\widehat{C}_k$ is the set of non-zero entries of an estimated column $\widehat{W}_{\cdot k}$ and $\widehat{K}$ is the estimated factor dimension. Then:
\begin{align}
\text{Relevance} &=\frac{1}{|\mathcal{C}_{f}|}\sum_{\widehat{C}_{k'} \in \mathcal{C}_f} \max_{C_{k} \in \mathcal{C}_t} J(C_k, \widehat{C}_{k'}) , \label{eq:rel}\\
\text{Recovery} &=  \frac{1}{|\mathcal{C}_t|}\sum_{C_k \in \mathcal{C}_t} \max_{\widehat{C}_{k'} \in \mathcal{C}_f}  J(C_k, \widehat{C}_{k'}),\label{eq:rec}\\
    \text{Consensus} &= \frac{1}{\max\{|\mathcal{C}_t|, |\mathcal{C}_f|\}} \sum_{C_k\in\mathcal{C}_t} J(C_k, \widehat{C}_{\pi(k)}), \label{eq:consensus}
\end{align}
where $\pi$ is the optimal assignment of the estimated $\widehat{C}_{k'}$ to the true $C_k$ (based on Jaccard scores), calculated using the Hungarian algorithm \citep{M57}.

The disentanglement score of \citet{eastwood2018framework} measures how important each estimated factor dimension $\widehat{\bm{z}}_{\cdot k'}$ is for a particular true factor dimension $\bm{z}_{\cdot k}$, penalizing estimated factors which are predictive of multiple true factor dimensions. The disentanglement score is calculated as follows.
\begin{enumerate}
    \item For each true factor dimension $k$, train a gradient boosted tree to predict ${z}_{i k}$ from $\widehat{\bm{z}}_i$.
    \item Record the importance of each $\widehat{{z}}_{i k'}$ for predicting ${z}_{i k}$ as $R_{k',k}$ (these are the importance scores from the gradient boosted tree).
    \item Calculate the score of $\widehat{\bm{z}}_{\cdot k'}$ as:
    \begin{align*}
D(\widehat{\bm{z}}_{\cdot k'}) = 1 - H_K(P_{k'\cdot})
    \end{align*}
    where $H_K(P_{k'\cdot})=-\sum_{k=1}^K P_{k', k}\log_K(P_{k',k})$ is the entropy and $P_{k',k}=R_{k',k}/\sum_{k=1}^K R_{k',k}$ denotes the weighted importance of $\widehat{z}_{ik'}$ for predicting $z_{ik}$. If $\widehat{z}_{ik'}$ is important for predicting a single  factor dimension, the score is 1. If $\widehat{z}_{ik}$ is equally important for all factor dimensions, the score is zero.
    \item The overall disentanglement score is the weighted sum:
    \begin{align}
D(\widehat{\bm{z}}) = \sum_{k'=1}^{\widehat{K}} \rho_{k'} D(\widehat{\bm{z}}_{\cdot k'}) \label{eq:dis-score}
    \end{align}
    where $\rho_{k'} = \sum_{k=1}^{K} R_{k',k}/\sum_{k,k'=1}^{K,\widehat{K}} R_{k',k}$. If $\widehat{z}_{ k'}$ is irrelevant for every  $\{z_k\}_{k=1}^K$, then $\rho_{k'}\approx 0$ and $\widehat{{z}}_{ k'}$ does not contribute to the disentanglement score. 
\end{enumerate}

\subsection{Experiment settings}

\subsubsection{Hyperparameter settings}

\Cref{tab:hyperparam} displays the fixed hyperparameter settings for each experiment.

\begin{table}[h]
    \centering
    \caption{Hyperparameter settings for experiments.}
    \label{tab:hyperparam}
    \begin{tabular}{lccc}
        \toprule
        \textbf{Hyperparameter} & \textbf{Gaussian} & \textbf{Synthetic RNA} & \textbf{Platelet} \\
        \midrule
        Batch Size & 512 & 512 & 512 \\
        Epochs & 500 & 800 & 800 \\
        Hidden Dimensions & [50, 50] & [256, 128] &  [256, 128] \\
        Initial $K_S$ & 30 & 50 & 50 \\
        Initial $K_m$ & 5 & 10 & 10 \\
        Learning Rate (not $W$) & $0.01$ &$0.001$ & $0.001$ \\
        Learning Rate ($W$) & $0.01$& $0.01$ & $0.01$ \\
        \bottomrule
    \end{tabular}
\end{table}

\parhead{Spike-and-slab lasso parameters.}
For the regularization parameters, we set $\lambda_1=0.1$. For $\lambda_0$, we adopt an annealing strategy. Specifically, for the first 5\% of epochs, $\lambda_0=1$. For the next 25\% of epochs, $\lambda_0=10$. For the remaining epochs, $\lambda_0=15$.

For all experiments, the MSSVAE takes the $\eta_k$ prior hyperparameters to be $a=1, b=G$, where $G$ is the number of observed features.

\parhead{Noise parameter settings.}
For the Gaussian MSSVAE, the prior on the noise variance is 
\begin{align}
\sigma_j^2 \sim \text{Inverse-Gamma}(\alpha, \beta). 
\end{align}
Following \citet{moran2022identifiable}, we set $\alpha=1.5$. The hyperparameter $\beta$ is set to a data-dependent value. Specifically, we first calculate the sample variance of each feature, $\bm{x}_{\cdot j}$. Then, we set $\beta$ such that the 5\% quantile of the sample variances is the 90\% quantile of the Inverse-Gamma prior.

\subsubsection{Parameter initializations}

\textbf{Noise parameter initializations.}
For the Gaussian MSSVAE, the variance $\sigma^2$ is set to the 0.05-quantile of the empirical variance of each of the $G$ columns \citep{moran2022identifiable}.

For the NB-MSSVAE, the inverse dispersion $\phi_g$ is initialized based on a moment estimator:
\begin{align*}
\phi_g^{(\mathrm{init})} = \frac{\widehat{\mu}_g^2}{\widehat{\sigma}^2_g - \widehat{\mu}_g},
\end{align*}
where $\widehat{\mu}_g$ and $\widehat{\sigma}^2_g$ are the empirical mean and variance, respectively, of gene $g$. 

$\bm{W}$ \textbf{initialization.} We initialize the entries of $\bm{W}^{(S)}$ and $\bm{W}^{(m)}$ to 1. 

\textbf{Neural network parameters.} We use the default PyTorch initialization.

\subsection{Additional results}\label{sec:add-res}
\subsubsection{Nonlinear Gaussian Factor Analysis}\label{sec:add-gauss-result}
Here we show the worst performing replicate in the Gaussian nonlinear factor analysis setting (\Cref{sec:gaussian-sim}). 

\begin{figure}
\centering
    \includegraphics[width=0.5\textwidth]{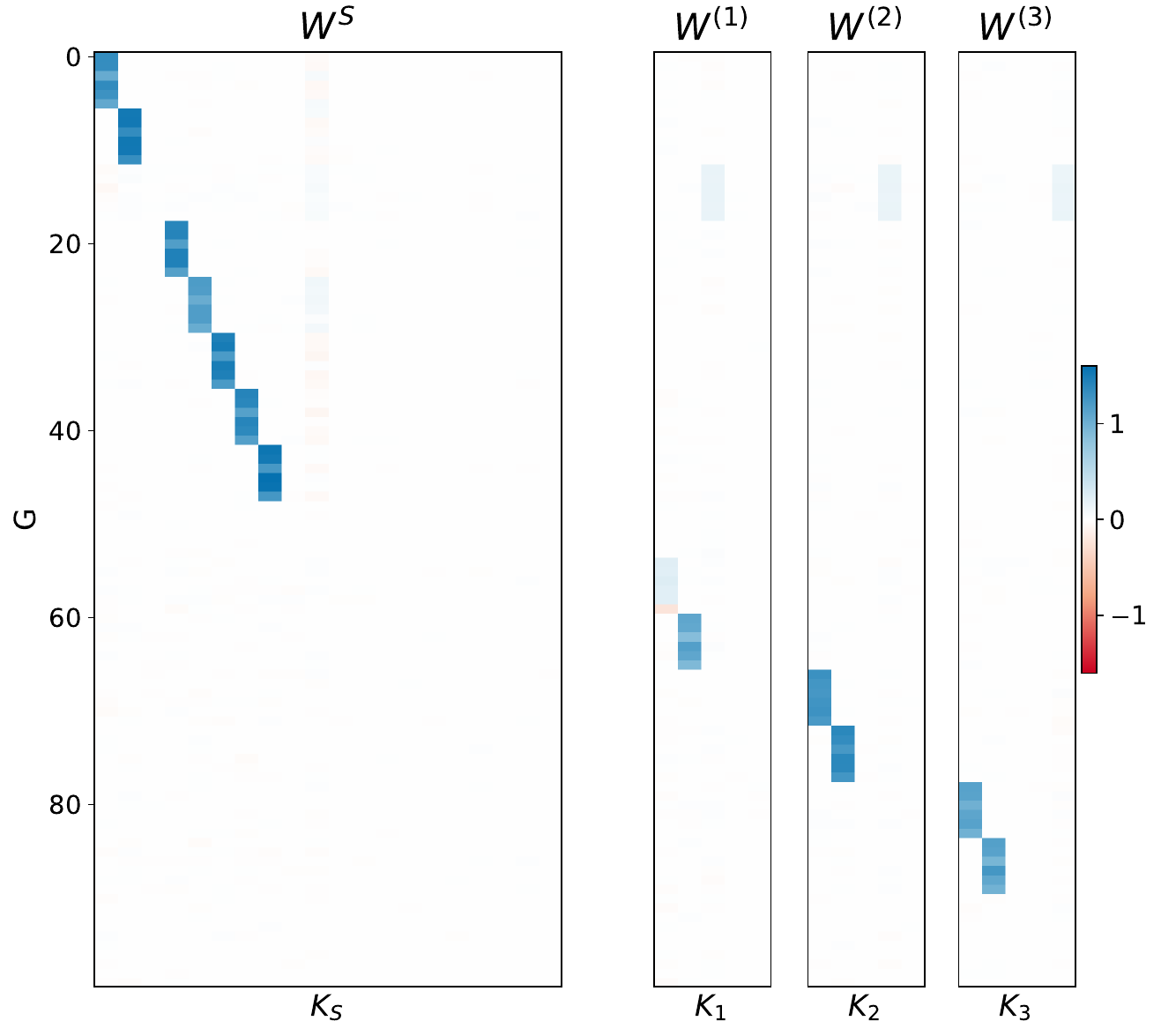}
    \caption{Gaussian nonlinear factor analysis: worst performing MSSVAE experiment in terms of average consensus score.} \label{fig:gauss_worst}
\end{figure}

\Cref{fig:gauss_worst} shows the estimated masking matrices for this data replicate; one of the shared blocks has instead been estimated as part of each of the study-specific blocks. This could be resolved through post-processing. 

\subsubsection{Synthetic Gene Expression Data}

\begin{figure}
\centering
\begin{minipage}[b]{0.49\textwidth}
\centering
\footnotesize\textbf{(a)}\par
\includegraphics[width=0.9\textwidth]{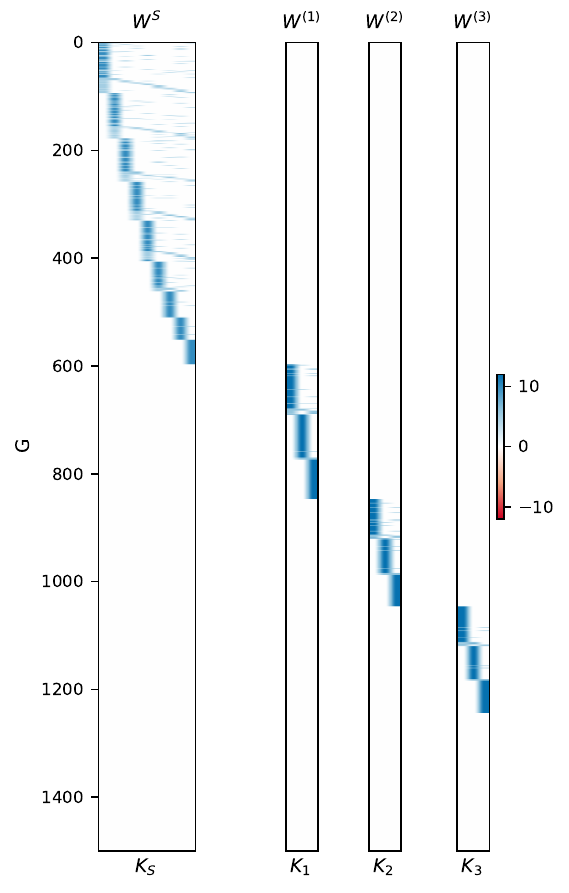}
\end{minipage}
\hfill
\begin{minipage}[b]{0.49\textwidth}
\centering
\footnotesize\textbf{(b)}\par
\includegraphics[width=0.9\textwidth]{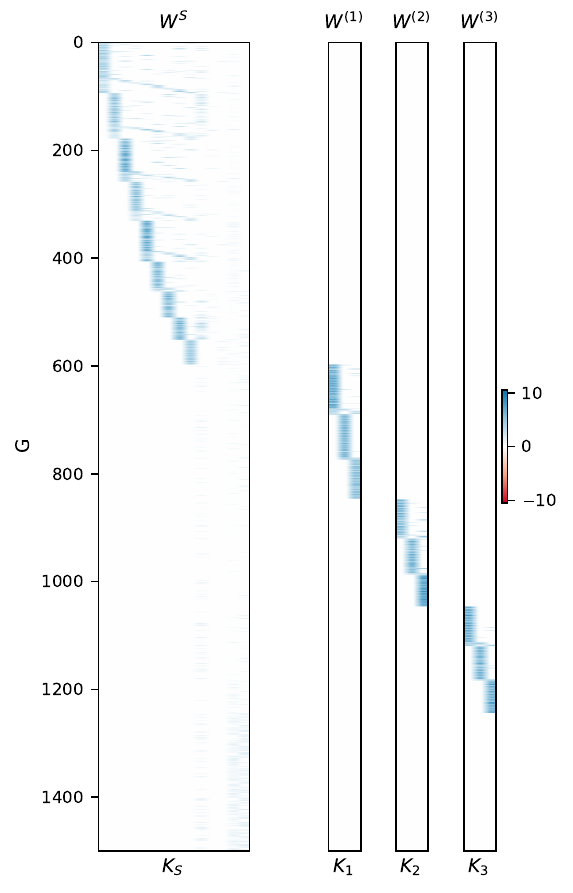}
\end{minipage}
\caption{Synthetic gene expression data: masking matrices for one of the 30 replicates. (a) True $\bm{W}^S, \bm{W}^{(m)}$. (b) Estimated $\widehat{\bm{W}}^{(S)}, \widehat{\bm{W}}^{(m)}$. Note we only show 1500 out of 5000 genes.}
\label{fig:syn-gene-W}
\end{figure}

\parhead{Data generation.} First, the base expression for gene $j$ in sample $i$ of group $m$, is generated from latent factors as:
\begin{flalign*}
\mu_{ij}^{(m)} &=
\begin{cases}
b_j + \sum_{k=1}^{K_S} W_{jk}^{(S)} z_{ik}^{(m)} + \sum_{k=1}^{K_m} W_{jk}^{(m)} \zeta_{ik}^{(m)} &\text{if } j \not\equiv 0 \mod 3\\
b_j + \sum_{k=1}^{K_S} W_{jk}^{(S)} \sin(3\pi/4\cdot z_{ik}^{(m)}) + \sum_{k=1}^{K_m} W_{jk}^{(m)} \sin(3\pi/4 \cdot \zeta_{ik}^{(m)})  &\text{if } j \equiv 0 \mod 3
\end{cases}
\end{flalign*}
where $b_j \sim N(0, 0.5^2)$ is a gene-specific intercept. The matrices $\bm{W}^{(S)}, \{\bm{W}^{(m)}\}_{m=1}^M$ have a mostly block structure with 5\% non-block, non-zero elements (see \Cref{fig:syn-gene-W}(a) for an example). The non-zero elements are drawn as ${w}_{jk}^{(S)} \stackrel{iid}{\sim} N(10, 0.01)$ and ${w}_{jk}^{(m)}\stackrel{iid}{\sim}N(12,0.01)$.

To ensure base expression is positive, we apply a softplus function. Then, we normalize so the gene expression levels for each sample sum to one:
\begin{align*}
{\mu}_{ij}^{(m)'} = \frac{\log(1+\exp(\mu_{ij}^{(m)}))}{\sum_{j=1}^G \log(1+\exp(\mu_{ij}^{(m)}))}.
\end{align*}
Note that this normalization induces correlation among all genes, making it more difficult to learn the true $\bm{W}$ matrices.

Next, the library size for each sample $i$ is generated as $\ell_i \sim \text{LogNormal}(12, 0.5^2)$. The sample mean is then calculated as $\widetilde{\mu}_{ij}^{(m)}=\ell_i \mu_{ij}^{(m)'}$.

To model the mean-variance trend in RNA-sequencing data \citep{mccarthy2012differential}, we follow \citet{zappia2017splatter} and simulate the biological coefficient of variation for each gene from a scaled inverse chi-squared distribution:
\begin{align*}
\text{BCV}_{ij}^{(m)} &= \left(\phi + 1/\sqrt{\widetilde{\mu}_{ij}^{(m)}} \right) \sqrt{\nu / q_{ij}^{(m)}}, \quad q_{ij}^{(m)} \sim \chi^2_\nu,
\end{align*}
with $\phi=0.1$ and $\nu=60$.

Finally, the observed data is generated as:
\begin{align*}
X_{ij}^{(m)} \sim \text{NB}(\text{mean}=\widetilde{\mu}_{ij}^{(m)}, \text{inv-disp}=1/(\text{BCV}_{ij}^{(m)})^2).
\end{align*}

\parhead{Results.} Here we provide \Cref{tab:comparison} with summary statistics comparing MSSVAE and multiGroupVI \citep{weinberger2022disentangling}. (This is the numeric summary of \Cref{fig:syn_gene_comparison}). \Cref{fig:syn-gene-W} shows the true and estimated masking matrices for one of the 30 replicates.

\begin{table}[!h]
\centering
\resizebox{\ifdim\width>\linewidth\linewidth\else\width\fi}{!}{
\begin{tabular}{lllllllllll}
\toprule
\multicolumn{1}{c}{ } & \multicolumn{2}{c}{DCI} & \multicolumn{2}{c}{AUPRC Shared} & \multicolumn{2}{c}{AUPRC Group 1} & \multicolumn{2}{c}{AUPRC Group 2} & \multicolumn{2}{c}{AUPRC Group 3} \\
\cmidrule(l{3pt}r{3pt}){2-3} \cmidrule(l{3pt}r{3pt}){4-5} \cmidrule(l{3pt}r{3pt}){6-7} \cmidrule(l{3pt}r{3pt}){8-9} \cmidrule(l{3pt}r{3pt}){10-11}
Method & Mean (SD) & Median & Mean (SD) & Median & Mean (SD) & Median & Mean (SD) & Median & Mean (SD) & Median\\
\midrule
NB-MSSVAE & 0.797 (0.028) & 0.799 & 0.979 (0.021) & 0.986 & 0.997 (0.014) & 1.000 & 0.993 (0.021) & 1.000 & 0.997 (0.011) & 1.000\\
multiGroupVI & 0.564 (0.023) & 0.565 & 0.972 (0.046) & 0.997 & 0.998 (0.006) & 1.000 & 0.997 (0.010) & 1.000 & 0.997 (0.009) & 1.000\\
\bottomrule
\end{tabular}}

\caption{Synthetic Gene Data: Comparison of MSSVAE and multiGroupVI.}\label{tab:comparison}
\end{table}

\subsection{Platelet data processing}

We process the platelet RNA-sequencing data as follows. 
\begin{itemize}
    \item Remove samples with total number of transcripts less than $e^{12.5}$;
    \item Keep only genes that had at least 25 counts over all samples;
    \item Keep only genes with counts-per-million greater than 1 for at least 20\% of samples;
    \item Keep the top 5000 most variable genes based on Pearson residuals of a negative binomial offset model \citep{lause2021analytic}, implemented using scanpy \citep{wolf2018scanpy}.
\end{itemize}

\subsection{Platelet results}

Here, we display the results of gene ontology enrichment analysis from \Cref{sec:platelet}.

\begin{figure}
    \includegraphics[width=\textwidth]{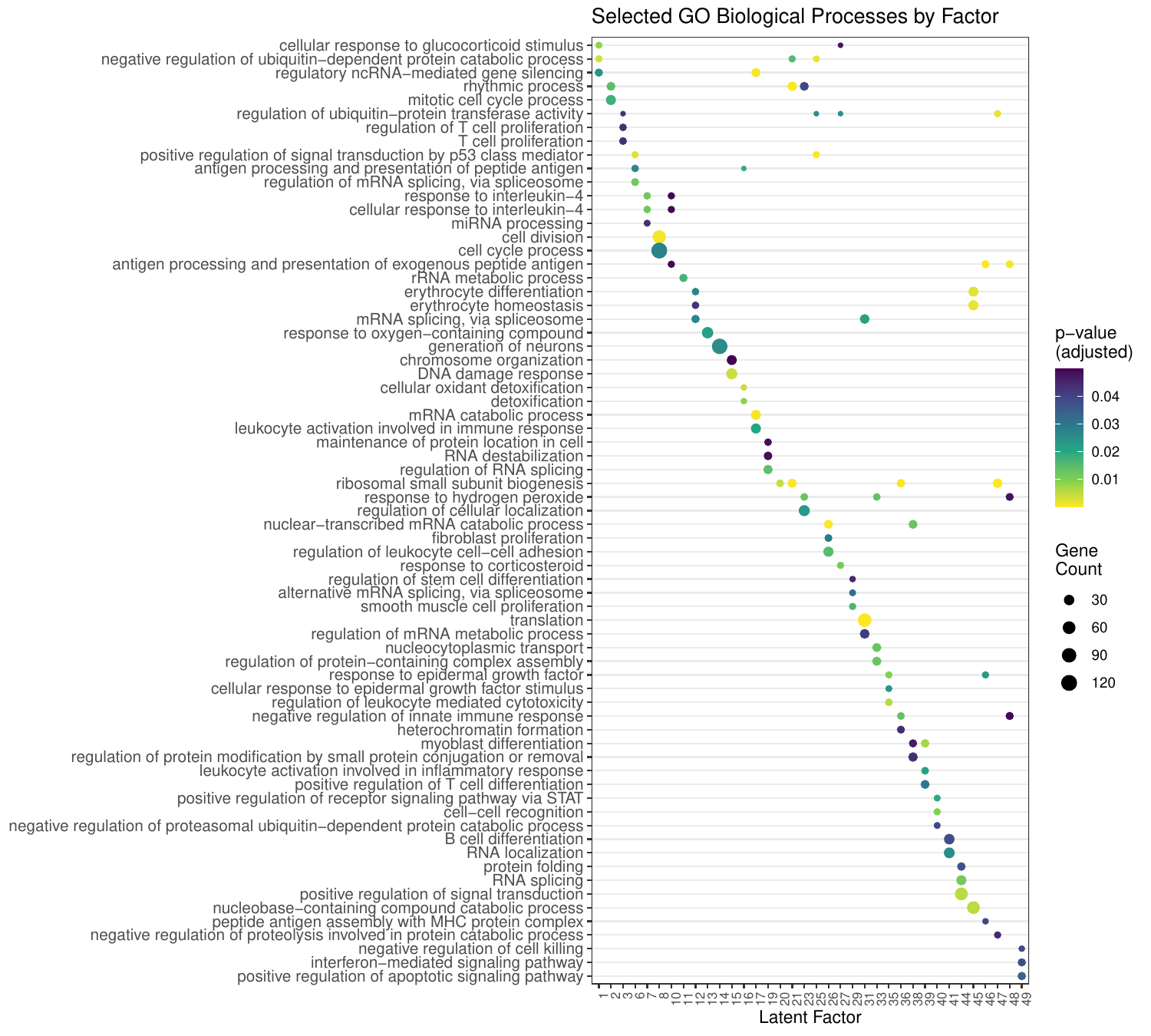}
    \caption{Summary of gene over-representation results for shared latent dimensions. For each latent dimension, we plot the top 3 biological processes by fold enrichment, after filtering for biological processes occurring in more than 10\% of latent dimensions. }\label{fig:platelet-shared}
\end{figure}

\begin{figure}
    \includegraphics[width=\textwidth]{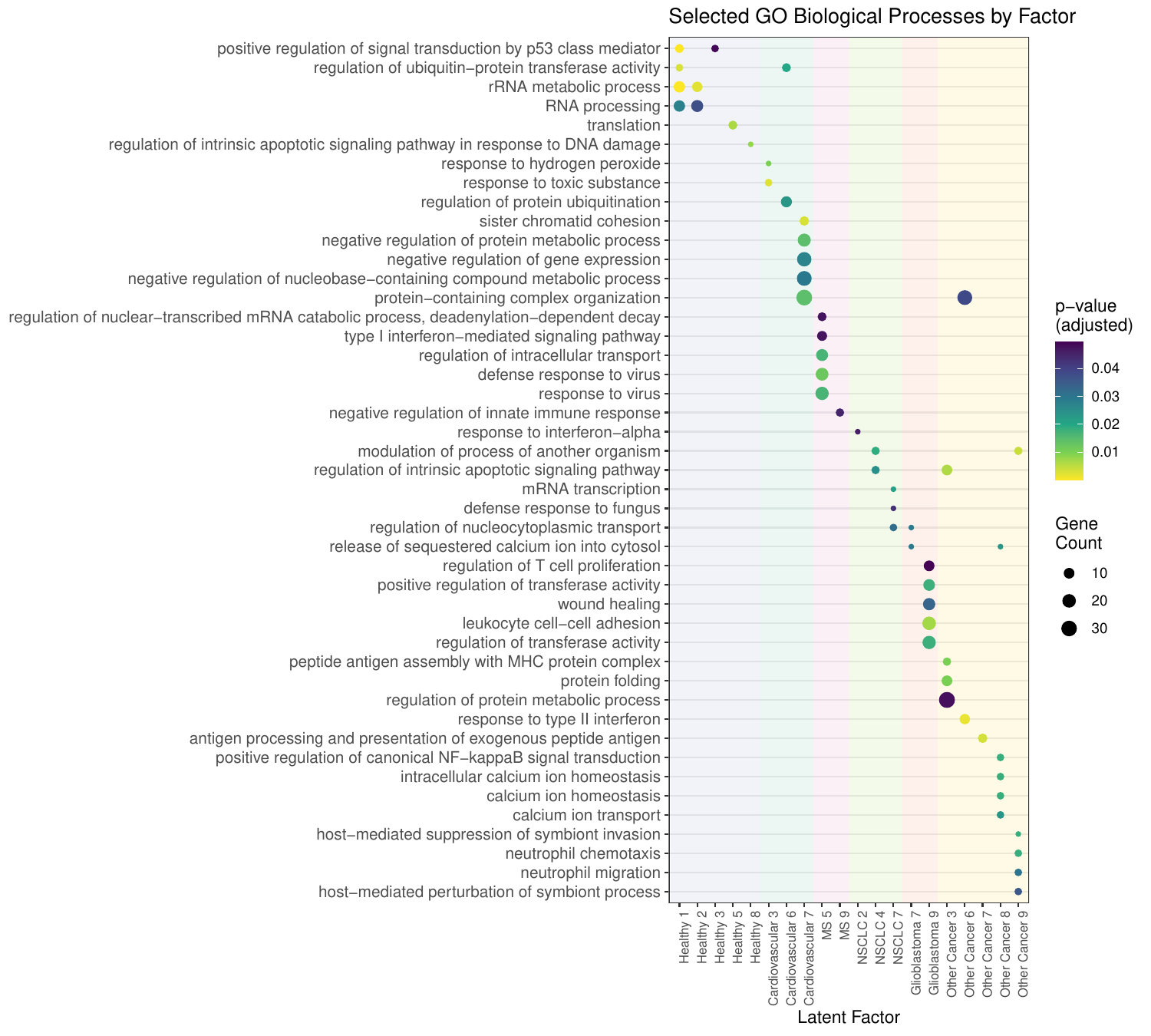}
    \caption{Summary of gene over-representation results for disease-state-specific latent dimensions. Displayed are the top 5 biological processes for each latent dimension by fold enrichment.}\label{fig:platelet-specific}
\end{figure}

\end{document}